\documentclass[10pt,twocolumn,letterpaper]{article}

\usepackage{cvpr}      

\usepackage{amsthm}
\usepackage{algorithm}
\usepackage{algpseudocode}
\newtheorem{proposition}{Proposition}

\makeatletter
\def\cvprsection{\@startsection {section}{1}{\z@}{-6pt plus -1pt minus -1pt}{3pt}{\large\bf}}
\def\cvprsubsection{\@startsection {subsection}{2}{\z@}{-5pt plus -1pt minus -1pt}{2pt}{\elvbf}}
\renewcommand\paragraph{\@startsection{paragraph}{4}{\z@}{2pt plus 1pt minus 1pt}{-1em}{\normalfont\normalsize\bfseries}}
\g@addto@macro\normalsize{%
  \setlength\abovedisplayskip{4pt plus 2pt minus 2pt}%
  \setlength\belowdisplayskip{4pt plus 2pt minus 2pt}%
  \setlength\abovedisplayshortskip{0pt plus 2pt}%
  \setlength\belowdisplayshortskip{2pt plus 2pt minus 1pt}}
\makeatother
\definecolor{cvprblue}{rgb}{0.21,0.49,0.74}
\usepackage[pagebackref,breaklinks,colorlinks,allcolors=cvprblue]{hyperref}

\def\confName{CVPR}
\def\confYear{2027}

\title{Target-Consistent Action Guidance via Semi-Unbalanced Optimal Transport for Drifting World Models}

\author{
Md Abtahi Majeed Chowdhury$^{1*}$,
Md Rifat Ur Rahman$^{1*}$,
Md Sakib Hossain Shovon$^{2}$, \\
Jaesik Choi$^{2}$,
Minhyuk Sung$^{2}$ \\[0.5em]
$^{1}$Bangladesh University of Engineering and Technology, \;
$^{2}$KAIST \\
{\small $^{*}$Equal contribution}
}

\begin{document}
\maketitle
\begin{abstract}
Action-conditioned world models let robots imagine the outcomes of their actions for planning and offline policy evaluation, but these uses need many rollouts, which multi-step diffusion makes expensive.
Drifting models move the iteration from inference to training and generate samples in one network pass.
DriftWorld~\cite{lu2026driftworld} brought this paradigm to action-conditioned world models; to strengthen action following, it repels predictions from the current frame, a construction borrowed from classifier-free guidance.
We show that this accentuation has a \emph{target bias}: the push persists even when a prediction matches the recorded future, in kernel, balanced, and unbalanced Wasserstein drifting fields alike.
We introduce Mass-Share Accentuation (MSA) to remove this bias.
MSA treats the current frame as a counterfactual destination competing with the recorded future in a semi-relaxed optimal transport problem; subtracting the closed-form share of the correct future makes the guidance vanish at the target, and the remainder is shared by all predictions.
We prove that the recorded future is the unique fixed point along the motion direction for collapsed predictions, and that the guidance fades with prediction error.
On Push-T, Robomimic, and BridgeV2, where trained models under-react rather than overshoot, MSA improves action ranking, response slope, and moving-region error over latent-space baselines at no inference cost, though these benchmarks do not isolate bias removal as the cause.
\end{abstract}w
\section{Introduction}
\label{sec:intro}

Action-conditioned world models predict how an environment changes in response to an agent's actions, so that a robot can imagine the outcome of a decision before executing it~\cite{ha2018world,hafner2020dream}.
Built on large-scale video generation, they now simulate fine-grained robot--object interaction~\cite{yang2024unisim,zhu2025irasim,guo2026ctrlworld} and long-horizon plans~\cite{du2024vlp,bruce2024genie,nvidia2025cosmos}.
They are used to rank candidate actions inside a control loop~\cite{hansen2022tdmpc,qi2026gpc} and to evaluate policies offline without hardware trials~\cite{quevedo2025worldgym,team2025veo}.
Both uses need many rollouts, yet most world models are diffusion-based~\cite{ho2020ddpm,ho2022vdm} and spend many network evaluations on every frame.

Drifting models~\cite{deng2026drifting} remove this cost by moving the iteration from inference to training: a drifting field attracts generated samples toward data and repels them from each other, and the one-step generator is trained until the field vanishes.
Follow-up work relates drifting to score matching~\cite{lai2026unified,turan2026secretly} and to Wasserstein gradient flows~\cite{he2026sinkhorndrifting,cao2026gradientflowdrifting}.
W-Flow~\cite{han2026wflow} derives the field as the Wasserstein gradient of a debiased Sinkhorn divergence, so that, under suitable assumptions, it vanishes only at the target, and UW-Flow~\cite{jeon2026uwflow} relaxes its marginal constraints with unbalanced transport.
Both were developed for class-conditional image generation.

\begin{figure}[t]
\centering
\includegraphics[width=\linewidth]{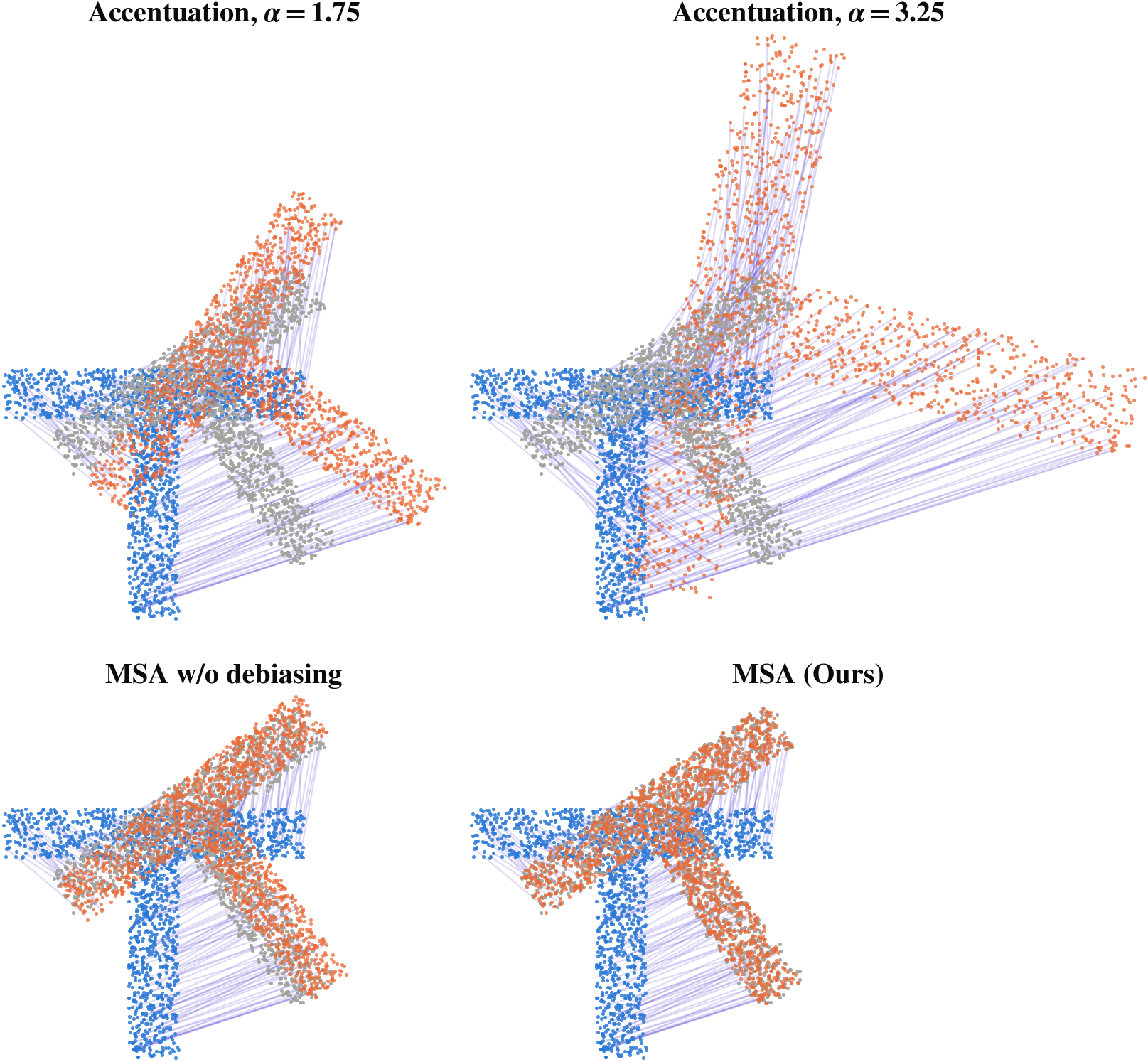}
\caption{\textbf{In a controlled toy, accentuation overshoots the recorded future.} Each point of a block (blue) is moved by an action to its recorded future (gray); orange shows a trained one-step world model's predictions and purple lines their motion. Accentuation drives predictions past the target, increasingly with $\alpha$. Without debiasing, a small overshoot remains; MSA lands on the recorded future. See Sec.~\ref{sec:exp:alpha}.}
\label{fig:hero}
\vspace{-10pt}
\end{figure}

DriftWorld~\cite{lu2026driftworld} brought drifting to action-conditioned world models with three adaptations: a conditional drifting field with a single positive, the recorded future, and an \emph{action-accentuated} negative distribution; a drifting loss in a DINOv2/v3 feature space~\cite{oquab2023dinov2,simeoni2025dinov3}; and a U-Net that conditions each frame on its own action.
To make the model follow the action rather than copy the current frame, DriftWorld also \emph{accentuates} the action by pushing predictions away from that frame.

Accentuation is borrowed from classifier-free guidance~\cite{ho2022cfg}, and it inherits guidance's extrapolation: the push does not stop at the recorded future.
In image generation this overshoot strengthens a class, but a world model has only one correct future, and a push that persists there drives the accentuated predictions toward exaggerated motion.
We call this the \emph{target bias} of accentuation.
It persists when accentuation is carried over to W-Flow, where every prediction receives the same constant push even when it is already correct, and unbalanced transport only weakens it.
Removing accentuation is no remedy, since the model then tends to copy the current frame~\cite{lu2026driftworld}.
The guidance should instead push a prediction only by the motion it is still missing.
This is a property of the field; trained world models on our benchmarks show the opposite failure, under-reacting to the action, so we evaluate removing target bias and improving action response separately.

We propose \emph{Mass-Share Accentuation} (MSA), which corrects this bias.
MSA treats the current frame as a competing destination rather than a negative: each prediction splits its mass between the recorded future and the current frame in one semi-relaxed transport problem, and the share sent to the current frame measures how much motion is missing.
Subtracting the share that the correct future itself would send makes the guidance vanish at the target, and applying it as one shared translation leaves the W-Flow field otherwise unchanged.
Our contributions are as follows.
\begin{itemize}
  \item We show that action accentuation keeps pushing predictions at the recorded future in kernel, balanced, and unbalanced drifting fields, and a controlled toy confirms the overshoot of kernel and balanced accentuation (\cref{sec:method:bias,sec:exp:alpha}).
  \item We introduce MSA, a bias-corrected accentuation built on counterfactual semi-relaxed transport with a closed-form assignment (\cref{sec:method:msa,sec:method:guidance}).
  \item We prove that the MSA field has the recorded future as its unique fixed point along the motion direction for collapsed predictions, and, for any predictions, adds no relative deformation to the base field and is bounded by the prediction error (\cref{sec:method:guidance}).
  \item On Push-T, Robomimic, and BridgeV2, MSA improves action response and motion fidelity over latent-space DriftWorld and W-Flow baselines at no inference cost; the bias removal itself is verified in a controlled setting (\cref{sec:experiments}).
\end{itemize}

\section{Related Work}
\label{sec:related}

\paragraph{Action-conditioned world models.}
Predicting future observations from actions has a long history in model-based reinforcement learning~\cite{ha2018world,hafner2020dream}.
Recent models instead adapt large video generators.
UniSim, GAIA-1, Genie, and Cosmos treat the world as a video that unrolls under actions or text~\cite{yang2024unisim,hu2023gaia,bruce2024genie,nvidia2025cosmos}.
Other work turns pretrained video diffusion models into interactive simulators~\cite{rigter2025avid,huang2026vid2world}, and robot-specific models such as IRASim, Ctrl-World, and UWM produce fine-grained, multi-view manipulation rollouts~\cite{zhu2025irasim,guo2026ctrlworld,zhu2025uwm}.
Nearly all of these models sample with multi-step diffusion or autoregression.
DriftWorld~\cite{lu2026driftworld} is the one-step exception and our starting point.
We keep its architecture and change only how the training field encourages the model to follow the action.

\paragraph{One-step generation.}
Diffusion models~\cite{sohldickstein2015deep,ho2020ddpm,song2021sde} and flow matching~\cite{lipman2023flow,liu2023rectified,albergo2023stochastic,ma2024sit} generate high-quality samples by integrating a stochastic or ordinary differential equation over many steps.
Distillation compresses a pretrained multi-step teacher into a one-step student~\cite{salimans2022progressive,luo2023diffinstruct,yin2024dmd,sauer2024add,zhou2024sid}, but it requires the teacher and inherits its errors.
Consistency-style methods train few-step models from scratch by enforcing self-consistency along a predefined diffusion or flow trajectory~\cite{song2023consistency,song2024ict,lu2024scm,frans2024shortcut,geng2025meanflow,geng2025imf,boffi2025flowmap,zhou2025imm}, so their training signal still comes from a trajectory designed for iterative sampling.
Drifting~\cite{deng2026drifting} and its Wasserstein refinements~\cite{han2026wflow,jeon2026uwflow} instead define the training-time evolution of the generator directly, with neither a teacher nor a differential equation, and are competitive with multi-step models at one network evaluation.
This makes drifting a natural backbone for world models, where each decision needs many rollouts.


\section{Preliminaries}
\label{sec:prelim}

A generator $f_\theta$ maps noise $\epsilon\sim p_\epsilon$ to $x=f_\theta(\epsilon)$ with pushforward $q_\theta=(f_\theta)_\# p_\epsilon$; $p$ is the data distribution, positives $y^+\sim p$, and negatives $y^-\sim q_\theta$.
We write $\mathrm{sg}(\cdot)$ for stop-gradient and use the cost $c(x,y)=\tfrac12\|x-y\|_2^2$.

\paragraph{Drifting models.}
A drifting field $V_{p,q}(x)$ moves a generated sample $x$ and vanishes when $q=p$; GMD uses $V_{p,q}=V^+_p-V^-_q$, two kernel mean-shift vectors that attract $x$ toward positives and repel it from negatives.
The generator is trained by the fixed-point regression
\begin{equation}
  \mathcal{L}(\theta)=\mathbb{E}_{\epsilon}\Big\|f_\theta(\epsilon)-\mathrm{sg}\big(f_\theta(\epsilon)+V_{p,q_\theta}(f_\theta(\epsilon))\big)\Big\|_2^2 ,
  \label{eq:drift_loss}
\end{equation}
which evolves $q_\theta$ toward $p$ during training while inference remains one network pass.
For an energy $\mathcal{F}(q;p)$ minimized at $q=p$, the Wasserstein gradient flow~\cite{jordan1998variational,ambrosio2008gradient} moves samples along $V=-\nabla_x\frac{\delta\mathcal{F}}{\delta q}$; an Euler step of it is the update in \cref{eq:drift_loss}, so an energy determines a drifting field~\cite{han2026wflow}.

\paragraph{Transport-based fields: W-Flow and UW-Flow.}
The entropic optimal transport cost~\cite{cuturi2013sinkhorn,peyre2019computational} over couplings $\Pi(q,p)$ is
\begin{equation}
\mathrm{OT}_\varepsilon(q,p)=\min_{\pi\in\Pi(q,p)}\int c\,\mathrm{d}\pi+\varepsilon\,\mathrm{KL}(\pi\,\|\,q\otimes p).
  \label{eq:eot}
\end{equation}
W-Flow~\cite{han2026wflow} takes $\mathcal{F}$ to be the debiased Sinkhorn divergence~\cite{genevay2018sinkhorn,feydy2019interpolating} $S_\varepsilon(q,p)=\mathrm{OT}_\varepsilon(q,p)-\tfrac12\mathrm{OT}_\varepsilon(q,q)-\tfrac12\mathrm{OT}_\varepsilon(p,p)$, whose field is
\begin{equation}
  V_{p,q}(x)=T_{q,p}(x)-T_{q,q}(x),
  \label{eq:wflow_v}
\end{equation}
with $T_{q,\mu}(x)=\int y\,\pi_{q,\mu}(\mathrm{d}y\mid x)$ the barycentric projection of the optimal coupling; the self term uses an independent generated batch $q'$.
Under suitable assumptions it vanishes everywhere iff $q=p$~\cite{feydy2019interpolating,han2026wflow}, though on the support of $q$ alone it can also vanish elsewhere (e.g., at $q=\delta_{\bar y}$, $\bar y$ the mean of $p$).
UW-Flow~\cite{jeon2026uwflow} replaces the marginal constraints by penalties of weight $\lambda>0$~\cite{liero2018optimal,chizat2018scaling},
\begin{equation}
\begin{aligned}
  \mathrm{UOT}_{\varepsilon,\lambda}(q,p)=\inf_{\pi\geq 0}\;&\int c\,\mathrm{d}\pi+\lambda\,\mathrm{KL}(\pi_1\|q)+\lambda\,\mathrm{KL}(\pi_2\|p)\\
  &+\varepsilon\,\mathrm{KL}(\pi\,\|\,q\otimes p),
\end{aligned}
\label{eq:uot}
\end{equation}
with marginals $\pi_1,\pi_2$ ($\lambda\to\infty$ recovers \cref{eq:eot}), and debiases it as $S_\varepsilon$~\cite{sejourne2019unbalanced}, giving
\begin{equation}
  V^\lambda_{p,q}(x)=s^\lambda_{q,p}(x)\big(T^\lambda_{q,p}(x)-x\big)-s^\lambda_{q,q}(x)\big(T^\lambda_{q,q}(x)-x\big),
  \label{eq:uwflow_v}
\end{equation}
where $s^\lambda_{q,\mu}=\mathrm{d}\pi_1/\mathrm{d}q$ is the mass transported from $x$ ($s<1$ for expensive transport) and $T^\lambda_{q,\mu}$ the normalized barycentric projection.

\paragraph{Semi-relaxed transport.}
Keeping only the source marginal $a$ exact~\cite{rabin2014adaptive,chizat2018scaling}, with destination weights $b$ and costs $C$,
\begin{equation}
  \min_{\Pi\geq 0}\;\langle C,\Pi\rangle+\varepsilon\,\mathrm{KL}(\Pi\,\|\,a\otimes b)
  \quad\text{s.t.}\quad \Pi\mathbf{1}=a
  \label{eq:semi_relaxed}
\end{equation}
decouples over rows into $\Pi_{ij}=a_i\,b_j e^{-C_{ij}/\varepsilon}\big/\sum_k b_k e^{-C_{ik}/\varepsilon}$: the destination masses follow from the costs.

\section{Method}
\label{sec:method}

\subsection{Setup}
\label{sec:method:setup}

Following DriftWorld~\cite{lu2026driftworld}, let $o_t$ and $a_t$ be the observation and action at time $t$.
The world model maps noise $\epsilon$ and the condition $c=(o_{t-F:t},a_{t:t+T})$ to a predicted future $x=f_\theta(\epsilon,c)$.
For each condition, the positive is the recorded future $y^+=o_{t+1:t+T+1}$, the negatives are $N_{\mathrm{neg}}$ predictions $x_i$ with empirical distribution $q$, and $q'$ is an independent generated batch for the self term of W-Flow.
The current frame $o_t$, repeated over the horizon, is the real no-action sample, and
\begin{equation}
  d=y^+-o_t
  \label{eq:motion}
\end{equation}
is the recorded motion.
As in DriftWorld, the loss is computed in a feature space $\phi$ separately at each location $(h,w)$; within one location, $x_i$, $y^+$, $o_t$, and $d$ denote features.
With the single positive $\delta_{y^+}$, $T_{q,y^+}\equiv y^+$, so the W-Flow field of \cref{eq:wflow_v} becomes
\begin{equation}
  V_{0,i}=y^+-T_{q,q'}(x_i),
  \label{eq:base_field}
\end{equation}
which vanishes at $q=\delta_{y^+}$.

\subsection{Target bias of action accentuation}
\label{sec:method:bias}

DriftWorld accentuates action following by mixing real no-action samples into the negatives,
\begin{equation}
  \tilde q=(1-\gamma)\,q_\theta+\gamma\,p(\cdot\mid\varnothing,o_{t-F:t}),
  \qquad \alpha=\tfrac{1}{1-\gamma}\geq 1,
  \label{eq:accentuation}
\end{equation}
the classifier-free-guidance construction of Drifting~\cite{deng2026drifting}.
Setting $\tilde q$ equal to the positives formally yields the guided target $\alpha\,p(\cdot\mid a_t,o_{t-F:t})-(\alpha-1)\,p(\cdot\mid\varnothing,o_{t-F:t})$, but with a single recorded future this is a signed measure, not a distribution.
We instead check whether the recorded future is stationary under DriftWorld's mean-shift field $V_{p,\tilde q}=V^+_p-V^-_{\tilde q}$, with
\begin{equation}
  V^-_{\tilde q}(x)=\frac{\mathbb{E}_{y^-\sim\tilde q}\big[k(x,y^-)(y^--x)\big]}{\mathbb{E}_{y^-\sim\tilde q}\big[k(x,y^-)\big]}
  \label{eq:meanshift}
\end{equation}
and $V^+_p(x)=y^+-x$ (we omit the implementation's extra normalizations).

\begin{proposition}[Residual push and overshoot]
\label{prop:dw_push}
Let $k(x,x)=1$, $\gamma>0$, and $\omega(x)=\frac{\gamma\kappa(x)}{(1-\gamma)+\gamma\kappa(x)}$ with $\kappa(x):=k(x,o_t)$.
(i)~If $q_\theta=\delta_{y^+}$ and $\kappa(y^+)>0$, then
\begin{equation}
  V_{p,\tilde q}(y^+)=\omega(y^+)\,d,\qquad \omega(y^+)\in(0,1).
  \label{eq:dw_push}
\end{equation}
(ii)~If $k$ is a positive, continuous, radially decreasing kernel vanishing at infinity, $d\neq 0$, and $q_\theta=\delta_x$ with $x=o_t+t\,d$, the field vanishes at some $t^\star>1$ satisfying $t^\star=1/\big(1-\omega(x^\star)\big)$.
\end{proposition}
\noindent The proof of Proposition 1 is in \textit{Supp.~A.1}.

A generator that already reproduces the recorded future is still pushed along the motion, and a collapsed one settles beyond it.
We call this the \emph{target bias} of accentuation; its size depends on the kernel affinities, whereas in Wasserstein fields it can be computed exactly.
Because DriftWorld conditions the network on $\alpha$ and samples it log-uniformly from $[1,4]$ during training (App.~C.2 of~\cite{lu2026driftworld}), the bias concerns outputs at $\alpha>1$; at $\alpha=1$, $\gamma=0$ and the field has no push, so any bias there can only arise through weights shared across $\alpha$, and degradation at $\alpha=1$ is not direct evidence of target bias.

\paragraph{Wasserstein accentuation.}
The analogue of \cref{eq:accentuation} in W-Flow transports the predictions to $y^+$ and, separately, to $o_t$, and adds $\alpha-1$ times the difference.
Both destinations are Dirac measures, so every prediction receives the same push,
\begin{equation}
  A^{\mathrm{bal}}_i=(\alpha-1)\big[(T_{q,y^+}(x_i)-x_i)-(T_{q,o_t}(x_i)-x_i)\big]=(\alpha-1)\,d .
  \label{eq:acc_bal}
\end{equation}
With unbalanced transport~\cref{eq:uwflow_v}, the push is weighted by the transported-mass ratios $s_{y^+}$, $s_{o_t}$:
\begin{equation}
  A^{\mathrm{UW}}_i=(\alpha-1)\big[s_{y^+}(x_i)\,d+\big(s_{y^+}(x_i)-s_{o_t}(x_i)\big)(o_t-x_i)\big].
  \label{eq:acc_uw}
\end{equation}

\begin{proposition}[Target bias of Wasserstein accentuation]
\label{prop:bal_bias}
(i)~For a collapsed $q=\delta_x$, $V_0+A^{\mathrm{bal}}$ vanishes only at $x^\star=y^++(\alpha-1)d$.
(ii)~At $q=\delta_{y^+}$, $s_{y^+}=1$, $s_{o_t}=\exp\!\big(-\|d\|^2/(2(2\lambda+\varepsilon))\big)$, and $A^{\mathrm{UW}}(y^+)=(\alpha-1)\,s_{o_t}\,d$, which is nonzero for $\alpha>1$ and $d\neq 0$.
\end{proposition}
\noindent The proof of Proposition 2 is in \textit{Supp.~A.2}.

Balanced accentuation moves the fixed point by $(\alpha-1)d$, and RMS normalization rescales the field without moving its zeros.
Unbalanced transport shrinks the push at the target but keeps it, and its second term in \cref{eq:acc_uw} deforms the generated cloud with a sample-dependent weight.
We therefore seek an accentuation term that
(D1)~keeps $y^+$ as the fixed point,
(D2)~pushes along $d$ while predictions lag and pulls back when they overshoot,
(D3)~adds no relative deformation to the base field, and
(D4)~lets every prediction contribute.

\subsection{Counterfactual semi-relaxed transport}
\label{sec:method:msa}

MSA treats $o_t$ not as a negative but as a second destination competing with $y^+$; Alg.~\ref{alg:msa} summarizes the full training step.
At one location, let $m^2=\|d\|^2$ be the squared motion magnitude and $M=m^2+\delta$, where a small offset $\delta>0$ keeps the costs finite for small motions.
We measure costs in these motion units, $C_{iv}=\|x_i-v\|^2/(2M)$ for $v\in\{y^+,o_t\}$, so that \emph{close} means the same at locations with large and small motion.
We solve the semi-relaxed problem \cref{eq:semi_relaxed} from the predictions to the two hypotheses,
\begin{equation}
  \min_{\Pi\geq 0}\;\langle C,\Pi\rangle+\varepsilon_J\,\mathrm{KL}(\Pi\,\|\,a\otimes b)
  \quad\text{s.t.}\quad \Pi\mathbf{1}=a,
  \label{eq:msa_ot}
\end{equation}
with $a_i=1/N_{\mathrm{neg}}$ and $b=(\tfrac12,\tfrac12)$.
The exact source marginal makes every prediction contribute (D4), and the free destination masses let the split between $y^+$ and $o_t$ express each prediction's preference for no motion.
Without the competing destination, a conserved-mass transport to $o_t$ alone would send all mass there and carry no signal.

\begin{proposition}[Closed-form allocation]
\label{prop:closed_form}
For $m>0$, let $r_i=\Pi_{i,o_t}/a_i$ be the fraction of the mass of prediction $i$ that the solution of \cref{eq:msa_ot} sends to $o_t$ (the rest goes to $y^+$). Then
\begin{equation}
  r_i=r(t_i)=\sigma\!\left(\frac{\tfrac12-t_i}{\varepsilon_d}\right),\quad
  t_i=\frac{\langle x_i-o_t,\,d\rangle}{m^2},\quad
  \varepsilon_d=\frac{\varepsilon_J M}{m^2},
  \label{eq:msa_closed_form}
\end{equation}
where $\sigma$ is the logistic sigmoid.
\end{proposition}
\noindent The proof of Proposition 3 is in \textit{Supp.~A.3}.

Here $t_i$ is the progress along the recorded motion ($0$ at $o_t$, $1$ at $y^+$), and $r_i$ decreases strictly with it (D2); errors orthogonal to $d$ do not affect it.
Motion-unit costs make $\varepsilon_J$ comparable across locations.
Since $\varepsilon_d=\varepsilon_J(1+\delta/m^2)$ grows as $m\to 0$, the gate flattens and the guidance fades on nearly static locations, and static locations ($m=0$) are skipped.
The shares are computed from detached predictions.

\begin{algorithm}[t]
\caption{MSA training step for one condition (Secs.~\ref{sec:method:msa}--\ref{sec:method:guidance}).}
\label{alg:msa}
\small
\begin{algorithmic}[1]
\Require condition $c$, recorded future $y^+$, current frame $o_t$, generator $f_\theta$, feature map $\phi$; $N_{\text{neg}}$, $\varepsilon_J$, $\delta$, $\chi$, $\beta$
\State $x_i \gets f_\theta(\epsilon_i, c)$,\; $x'_i \gets \mathrm{sg}(f_\theta(\epsilon'_i, c))$, \; $i=1,\dots,N_{\text{neg}}$ \Comment{two batches}
\Statex \textit{Lines 2--9 operate on the features $\phi(\cdot)$ of $x_i$, $x'_i$, $y^+$, $o_t$ at each location.}
\State $d \gets y^+ - o_t$;\; $V_{0,i} \gets y^+ - T_{q,q'}(x_i)$ \Comment{Eq.~(\ref{eq:base_field})}
\For{each location $(h,w)$ with $m^2=\|d_{h,w}\|^2>0$} \Comment{static: $A_{h,w}=0$}
  \State $t_i \gets \langle \mathrm{sg}(x_i)-o_t,\, d\rangle / m^2$;\; $\varepsilon_d \gets \varepsilon_J (m^2+\delta)/m^2$
  \State $r_i \gets \sigma\big((\tfrac12 - t_i)/\varepsilon_d\big)$;\; $r^+ \gets \sigma\big(-1/(2\varepsilon_d)\big)$ \Comment{Eq.~(\ref{eq:msa_closed_form})}
  \State $G_{h,w} \gets \frac{1}{N_{\text{neg}}}\sum_i r_i - r^+$;\; $A_{h,w} \gets \beta\, G_{h,w}\, d_{h,w}$ \Comment{Eq.~(\ref{eq:msa_guidance})}
\EndFor
\State $A \gets A \cdot \min\!\big(1,\, \chi\,\mathrm{RMS}(V_0)/\mathrm{RMS}(A)\big)$ \Comment{bounded share; skipped if $A=0$}
\State $V_i \gets V_{0,i} + A$;\; $F_i \gets V_i / \mathrm{RMS}(V)$ \Comment{same $A$ for all $i$}
\State $\mathcal{L} \gets \frac{1}{N_{\text{neg}}}\sum_i \big\| \phi(x_i) - \mathrm{sg}\big(\phi(x_i) + F_i\big) \big\|_2^2$ \Comment{Eq.~(\ref{eq:drift_loss}), summed over locations}
\State update $\theta$ with one optimizer step on $\mathcal{L}$
\Statex \textbf{Inference:} $x = f_\theta(\epsilon, c)$, one network pass; no $\alpha$, no extra solve.
\end{algorithmic}
\end{algorithm}

\subsection{Bias-corrected shared guidance}
\label{sec:method:guidance}

The share alone still has a floor at the target, $r^+=r(1)=\sigma(-1/(2\varepsilon_d))>0$, so $\beta\,r_i d$ would push a correct prediction, where $\beta>0$ is the guidance weight.
MSA subtracts this floor, averages over predictions, and applies one shared vector:
\begin{equation}
\begin{aligned}
  G&=\frac{1}{N_{\mathrm{neg}}}\sum_{i} r_i-r^+,\\
  A_{h,w}&=\beta\,G_{h,w}\,d_{h,w},\qquad V_i=V_{0,i}+A.
\end{aligned}
\label{eq:msa_guidance}
\end{equation}
The weight $\beta$ plays the role of $\alpha-1$ but is fixed and not a network input, so MSA needs no sampling-time scale.

\begin{proposition}[Bias correction]
\label{prop:bias_correction}
Let $m>0$ and $q=\delta_x$ with $x=o_t+t\,d+e$, $e\perp d$.
(i)~With the uncentered term $\beta\,r(t)d$, the field vanishes only at $e=0$, $t=\tilde t$ with $\tilde t-1=\beta\,r(\tilde t)\in\big(0,\beta\,r^+\big)$.
(ii)~With MSA, it vanishes only at $x=y^+$.
\end{proposition}
\noindent The proof of Proposition 4 is in \textit{Supp.~A.4}.

Subtracting $r^+$ thus removes the remaining bias for collapsed clouds, and at $q=\delta_{y^+}$ every $t_i=1$, so $G=0$ and $V=0$.
Since $r$ decreases, $G>0$ when every prediction lags ($t_i<1$) and $G<0$ when every prediction overshoots.
With spread the correction is not exact: $r$ is convex for $t>\tfrac12$, so for a cloud in this region with mean progress $\bar t=1$, Jensen's inequality gives
\begin{equation}
  G\approx\tfrac12\,r''(1)\,\mathrm{Var}(t)\geq 0,\qquad r''(1)=\frac{r^+(1-r^+)(1-2r^+)}{\varepsilon_d^2},
  \label{eq:msa_spread}
\end{equation}
a small forward push that vanishes as the cloud contracts.
Because $A$ is shared, $V_i-\bar V=V_{0,i}-\bar V_0$: MSA translates the cloud without deforming it (D3).

\begin{proposition}[Residual control]
\label{prop:residual}
At every location with $m>0$,
\begin{equation}
  \|G\,d\|\leq\frac{1}{4\varepsilon_d}\cdot\frac{1}{N_{\mathrm{neg}}}\sum_{i}\|x_i-y^+\|.
  \label{eq:msa_residual}
\end{equation}
\end{proposition}
\noindent The proof of Proposition 5 is in \textit{Supp.~A.5}.

The guidance therefore fades with the prediction error, unlike the constant push of \cref{eq:acc_bal}.

\paragraph{Training.}
Per condition, we compute $V_0$ with W-Flow's Sinkhorn solver and $A$ in closed form, rescale $A$ so that $\mathrm{RMS}(A)\leq\chi\,\mathrm{RMS}(V_0)$, normalize $F_i=V_i/\mathrm{RMS}(V)$, and regress $\phi(f_\theta(\epsilon_i))$ onto $\mathrm{sg}(\phi(x_i)+F_i)$ as in \cref{eq:drift_loss}, summed over locations.
Both rescalings multiply by positive factors; since $A$ and the $d$-component of $V_0$ have the same sign on either side of $t=1$, they keep the zero of \cref{prop:bias_correction} at $x=y^+$.
MSA adds no Sinkhorn solve and no sampling-time guidance; inference remains one network pass.

\paragraph{Discussion.}
Near the target, MSA reduces to the natural first-order correction: $\mathrm{d}r/\mathrm{d}t=-r(1-r)/\varepsilon_d$ gives
\begin{equation}
  G\approx c_d\,(1-\bar t),\qquad c_d=\frac{r^+(1-r^+)}{\varepsilon_d},
  \label{eq:msa_linear}
\end{equation}
with $\bar t$ the mean progress, that is, a shared projection of the mean residual $y^+-\bar x$ onto $d$ with gain $\beta c_d$.
MSA thus keeps the correct local behavior and differs from such a correction only away from the target.
This correction is the high-temperature limit of MSA ($\varepsilon_J\to\infty$,
$\beta = 4\varepsilon_J\lambda$); away from the target, the logistic gate raises the gain
above $c_d$ while predictions lag, and outperforms the projected-residual gate at the
matched local gain $\lambda = c_d$ under the same training (\textit{Supp.~E.2}).
Compared with DriftWorld, where $o_t$ enters a kernel-affinity pool of negatives, MSA compares $o_t$ directly with $y^+$, subtracts the correct future's share, and applies one shared translation; the two-class entropic softmax itself is standard.
Our analysis also assumes a single recorded future per condition; when futures are stochastic (e.g., BridgeV2), $y^+$ is one sample of the conditional distribution, and $G=0$ at $y^+$ does not imply zero guidance at that distribution.
These are properties of the training field, not guarantees on generalization or action ranking, and unit-RMS normalization may slow the correction of errors orthogonal to $d$; the variant $F_i=(V_{0,i}+A)/\mathrm{RMS}(V_0)$ avoids this at the cost of a larger step.

\section{Experiments}
\label{sec:experiments}
\setlength{\textfloatsep}{6pt plus 2pt minus 2pt}
\setlength{\dbltextfloatsep}{6pt plus 2pt minus 2pt}
\captionsetup[table]{skip=3pt}
\setlength{\floatsep}{6pt plus 2pt minus 2pt}
\setlength{\dblfloatsep}{6pt plus 2pt minus 2pt}

\begin{table*}[t]
\centering
\caption{\textbf{Results on Push-T and Robomimic.} Action: Kendall $\tau$, response slope (ideally 1) and correlation of the predicted object displacement across proposals, and sample noise (px / mm). Motion: moving-region MSE, static drift (frames 9--60), static MSE, and late-contact fraction at two thresholds. Standard: frame MSE, SSIM, PSNR, LPIPS. Best per dataset in \textbf{bold}.}
\label{tab:main}
\footnotesize
\setlength{\tabcolsep}{2.8pt}
\renewcommand{\arraystretch}{0.94}
\begin{tabular}{ll cccc ccccc cccc}
\toprule
& & \multicolumn{4}{c}{Action} & \multicolumn{5}{c}{Motion} & \multicolumn{4}{c}{Standard} \\
\cmidrule(lr){3-6}\cmidrule(lr){7-11}\cmidrule(lr){12-15}
& & & & & & & & & \shortstack{1\,px /\\2\,mm} & \shortstack{3\,px /\\5\,mm} & & & & \\
Dataset & Method & $\tau\uparrow$ & Slope$\uparrow$ & Corr.$\uparrow$ & Noise$\downarrow$ & Mov.$\downarrow$ & Drift$_{9\text{--}60}\downarrow$ & Static$\downarrow$ & Late$\downarrow$ & Late$\downarrow$ & MSE$\downarrow$ & SSIM$\uparrow$ & PSNR$\uparrow$ & LPIPS$\downarrow$ \\
\midrule
 & Kernel & 0.682 & 0.699 & 0.826 & 3.43 & 0.1806 & 0.0218 & 0.0192 & 0.42 & 0.45 & 0.0247 & 0.951 & 24.33 & 0.0607 \\
 & Kernel + acc. & 0.683 & 0.698 & 0.843 & 3.33 & 0.1846 & 0.0221 & 0.0195 & 0.39 & 0.42 & 0.0248 & 0.951 & 24.31 & 0.0602 \\
\raisebox{0pt}[0pt][0pt]{Push-T} & W-Flow & 0.693 & 0.713 & 0.830 & 1.49 & 0.1720 & 0.0208 & 0.0183 & 0.41 & 0.46 & 0.0244 & 0.952 & 24.48 & 0.0608 \\
 & W-Flow + acc. & 0.678 & 0.698 & 0.852 & \textbf{1.32} & 0.1761 & 0.0220 & 0.0194 & 0.42 & 0.42 & 0.0270 & 0.948 & 23.93 & 0.0663 \\
 & MSA (ours) & \textbf{0.712} & \textbf{0.746} & \textbf{0.856} & 1.55 & \textbf{0.1699} & \textbf{0.0205} & \textbf{0.0179} & \textbf{0.31} & \textbf{0.35} & \textbf{0.0240} & \textbf{0.954} & \textbf{24.64} & \textbf{0.0589} \\
\midrule
 & Kernel & 0.587 & 0.720 & 0.786 & 2.18 & 0.0847 & 0.0087 & \textbf{0.0123} & 0.57 & 0.43 & \textbf{0.0175} & 0.919 & 26.33 & 0.0409 \\
 & Kernel + acc. & 0.574 & 0.664 & 0.742 & 2.41 & 0.0872 & 0.0088 & 0.0132 & 0.44 & 0.39 & 0.0185 & 0.916 & 26.20 & 0.0418 \\
\raisebox{0pt}[0pt][0pt]{\begin{tabular}[c]{@{}l@{}}Robomimic\\Lift\end{tabular}} & W-Flow & 0.584 & 0.721 & 0.785 & 2.30 & 0.0839 & 0.0088 & 0.0124 & 0.58 & 0.37 & 0.0176 & 0.918 & 26.27 & \textbf{0.0404} \\
 & W-Flow + acc. & 0.563 & 0.713 & 0.748 & 2.53 & 0.0927 & 0.0087 & 0.0136 & 0.50 & 0.37 & 0.0192 & 0.914 & 25.96 & 0.0466 \\
 & MSA (ours) & \textbf{0.604} & \textbf{0.756} & \textbf{0.790} & \textbf{2.02} & \textbf{0.0788} & \textbf{0.0082} & 0.0129 & \textbf{0.39} & \textbf{0.30} & 0.0186 & \textbf{0.927} & \textbf{26.48} & 0.0413 \\
\midrule
 & Kernel & 0.670 & 0.776 & 0.777 & \textbf{4.44} & 0.1194 & 0.0052 & \textbf{0.0084} & 0.37 & 0.31 & 0.0133 & 0.939 & 26.70 & 0.0357 \\
 & Kernel + acc. & 0.660 & 0.769 & 0.781 & 4.89 & 0.1241 & 0.0050 & 0.0092 & 0.38 & 0.36 & 0.0143 & 0.937 & \textbf{26.74} & 0.0357 \\
\raisebox{0pt}[0pt][0pt]{\begin{tabular}[c]{@{}l@{}}Robomimic\\Can\end{tabular}} & W-Flow & 0.657 & 0.774 & 0.778 & 6.24 & 0.1203 & 0.0056 & 0.0087 & 0.31 & 0.31 & 0.0136 & 0.934 & 26.46 & 0.0375 \\
 & W-Flow + acc. & 0.654 & 0.795 & 0.780 & 6.50 & 0.1297 & 0.0054 & 0.0096 & 0.30 & 0.25 & 0.0148 & 0.927 & 26.21 & 0.0411 \\
 & MSA (ours) & \textbf{0.680} & \textbf{0.826} & \textbf{0.795} & 4.79 & \textbf{0.1181} & \textbf{0.0049} & 0.0086 & \textbf{0.27} & \textbf{0.23} & \textbf{0.0122} & \textbf{0.957} & 26.68 & \textbf{0.0352} \\
\bottomrule
\end{tabular}
\end{table*}

We evaluate MSA on three simulated manipulation benchmarks, Push-T~\cite{chi2023diffusionpolicy} (GPC data~\cite{qi2026gpc}) and Robomimic Lift and Can~\cite{mandlekar2021robomimic}, and on real robot video from BridgeV2~\cite{walke2023bridgedata}.

\subsection{Datasets and training setup}
\label{sec:exp:setup}

\paragraph{Shared setup.}
All methods share the one-step latent U-Net, recipe~\cite{lu2026driftworld}, data, steps, and noise seeds (\textit{Supp.~B.2}), and compute their fields on SD3 VAE latents~\cite{esser2024sd3}; the kernel baselines are thus a latent-space DriftWorld, not its published feature-space model (\textit{Supp.~B.3}), and our comparisons do not establish superiority over the original DriftWorld.

\paragraph{Baselines.}
\emph{Kernel} is the drifting field of DriftWorld and \emph{W-Flow}~\cite{han2026wflow} that of Eq.~\eqref{eq:base_field}; \emph{+ acc.} adds the accentuation of Eqs.~\eqref{eq:accentuation} and~\eqref{eq:acc_bal}, respectively, with the network conditioned on $\alpha\sim\log\mathcal{U}[1,4]$ and sampled at $\alpha=1$ (Sec.~\ref{sec:exp:alpha} varies $\alpha$).
\emph{MSA} adds Eq.~\eqref{eq:msa_guidance} to W-Flow with one fixed setting on every dataset and no $\alpha$ input.

\paragraph{Metrics.}
\emph{Action} metrics follow GPC~\cite{qi2026gpc}: $\tau$ is the Kendall correlation of predicted and simulated rewards over action proposals; slope and correlation compare predicted with simulated object displacements (slope below 1: under-reaction).
\emph{Motion} metrics score moving and static regions and motion onset (\textit{Supp.~B.4 and~B.5}).

\subsection{Results on Push-T and Robomimic}
\label{sec:exp:main}

In Tab.~\ref{tab:main} (rollouts in Fig.~\ref{fig:qual}), MSA has the best $\tau$, slope, and correlation on all three benchmarks.
Over the best baseline, $\tau$ rises by 0.019, 0.017, and 0.010 and the slope by 0.033, 0.035, and 0.031.
Each method is trained once per dataset, so the evidence is that MSA ranks first on three separately trained benchmarks, not one margin.
The accentuated baselines instead lower the $\tau$ of their base field in five of six cases and raise its moving-region MSE in all six; since they are sampled at $\alpha=1$, where the push vanishes (\cref{sec:method:bias}), this is not direct evidence of target bias.

All slopes are below 1: every model under-reacts rather than overshoots, so these benchmarks test action response, not target bias, and do not show that removing the bias causes MSA's gains.
MSA raises the slope and has the lowest moving-region MSE (by 1\% on Push-T and Can, 6\% on Lift, and 8\% over full episodes, Tab.~\ref{tab:fullep}), static drift, and late-contact fraction (0.39 vs.\ 0.44 or more on Lift) on every benchmark: the stronger response does not cost motion accuracy.

MSA is not best in noise on Push-T (1.55 vs.\ 1.32~px) and Can (4.79 vs.\ 4.44~mm), static MSE on Lift and Can, frame MSE (6\% above the best) and LPIPS on Lift, and PSNR on Can.

\subsection{Results on BridgeV2}
\label{sec:exp:bridge}

\emph{Controllability} ($\Delta$PSNR) follows Genie~\cite{bruce2024genie}; the \emph{motion probe} measures the progress of one-step samples along the real latent motion (1: full motion; \textit{Supp.~B.6}).

In Tab.~\ref{tab:bridge}, MSA is the best model in controllability and progress: its $\Delta$PSNR, measured on decoded frames, is 2.401 against at most 2.185, and it realizes 0.533 of the real latent motion against at most 0.443, at unchanged spread.
Its off-motion error (0.337) is second to W-Flow's (0.326).

Quality metrics do not test correctness: an unrelated real video beats every model in FID and FVD, and a motionless copy of the last frame beats every model in LPIPS and FID and every baseline in SSIM and FVD.
We therefore treat them as secondary to action and motion fidelity.
Among the models, MSA has the best SSIM, PSNR, LPIPS, and FVD, and is the only one that beats the copy in SSIM (.847 vs.\ .832) and FVD (243.1 vs.\ 251.7; baselines 304.4 or more); its FID is the highest (58.83 vs.\ 55.44 to 56.40), which we do not explain.

\subsection{Downstream planning with GPC-Rank}
\label{sec:exp:gpc}

\begin{table}[t]
\centering
\caption{\textbf{GPC-Rank on Push-T.} Each world model inside GPC-Rank~\cite{qi2026gpc}, two proposal policies. IoU: final T-block/goal overlap (mean $\pm$ s.d.\ over 50 episodes); s/plan: seconds per plan. Best in \textbf{bold}.}
\label{tab:gpc}
\footnotesize
\setlength{\tabcolsep}{2.5pt}
\begin{tabular}{l cc cc}
\toprule
& \multicolumn{2}{c}{Policy 1} & \multicolumn{2}{c}{Policy 2} \\
\cmidrule(lr){2-3}\cmidrule(lr){4-5}
Method & IoU$\uparrow$ & s/plan$\downarrow$ & IoU$\uparrow$ & s/plan$\downarrow$ \\
\midrule
Base policy & $0.597\pm0.163$ & -- & $0.644\pm0.138$ & -- \\
\midrule
Kernel & $0.704\pm0.188$ & 1.92 & $0.720\pm0.207$ & 1.96 \\
Kernel + acc. & $0.706\pm0.213$ & 1.84 & $0.721\pm0.166$ & 1.94 \\
W-Flow & $0.717\pm0.207$ & 1.88 & $0.722\pm0.169$ & 1.87 \\
W-Flow + acc. & $0.717\pm0.221$ & 1.79 & $0.721\pm0.173$ & \textbf{1.78} \\
MSA (ours) & $\mathbf{0.744}\pm0.200$ & \textbf{1.78} & $\mathbf{0.740}\pm0.159$ & 1.79 \\
\bottomrule
\end{tabular}
\end{table}

\begin{figure}[t]
\centering
\begin{subfigure}[t]{0.48\linewidth}
\centering
\includegraphics[width=\linewidth]{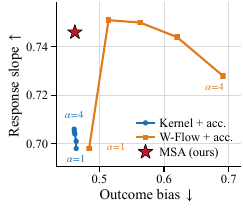}
\caption{}
\label{fig:alpha:a}
\end{subfigure}\hfill
\begin{subfigure}[t]{0.48\linewidth}
\centering
\includegraphics[width=\linewidth]{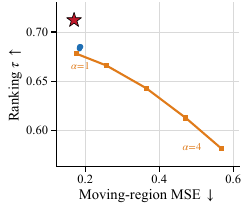}
\caption{}
\label{fig:alpha:b}
\end{subfigure}
\caption{\textbf{Accentuation scale sweep.} Curves trace $\alpha\in\{1,\dots,4\}$; MSA (no $\alpha$) is a star. (a) Slope$\uparrow$ vs.\ outcome bias$\downarrow$. (b) $\tau\uparrow$ vs.\ moving-region MSE$\downarrow$.}
\label{fig:alpha}
\end{figure}

\begin{table}[t]
\centering
\caption{\textbf{Target bias in the T-block toy} (Fig.~\ref{fig:hero}). Converged progress along the recorded motion (1 = recorded future) for accentuation scale $\alpha$ (for MSA, weight $\beta=\alpha-1$), and endpoint error at $\alpha=2.5$; mean over 3 seeds (s.d.\ $\le 0.001$). Balanced accentuation converges to $\bar t=\alpha$ (Prop.~\ref{prop:bal_bias}); MSA converges to the recorded future for every weight (Prop.~\ref{prop:bias_correction}). Best in \textbf{bold}.}
\label{tab:toy}
\footnotesize
\setlength{\tabcolsep}{3pt}
\begin{tabular}{l cccc c}
\toprule
& \multicolumn{4}{c}{$\bar t$ at $\alpha$} & Endpoint err.$\downarrow$ \\
\cmidrule(lr){2-5}
Method & 1.75 & 2.5 & 3.25 & 5 & ($\alpha=2.5$) \\
\midrule
W-Flow & \textbf{1.001} & \textbf{1.001} & \textbf{1.001} & \textbf{1.001} & \textbf{.003} \\
Kernel + acc. & 1.019 & 1.031 & 1.049 & 1.100 & .021 \\
UW-Flow + acc. & \textbf{1.001} & \textbf{1.001} & \textbf{1.001} & \textbf{1.001} & .004 \\
W-Flow + acc. & 1.754 & 2.504 & 3.253 & 5.004 & 1.050 \\
MSA w/o debiasing & 1.069 & 1.106 & 1.135 & 1.187 & .072 \\
MSA (ours) & \textbf{1.001} & \textbf{1.001} & \textbf{1.001} & \textbf{1.001} & \textbf{.003} \\
\bottomrule
\end{tabular}
\end{table}

\begin{figure*}[t]
\centering
\footnotesize
\setlength{\tabcolsep}{0pt}
\renewcommand{\arraystretch}{0}%
\newlength{\qualw}\setlength{\qualw}{0.089\textheight}
\newcommand{\qualimg}[1]{\parbox[c][\qualw][c]{\qualw}{\includegraphics[width=\qualw]{#1}}}
\newcommand{\qualph}[1]{\fcolorbox{gray!60}{gray!8}{\parbox[c][\dimexpr\qualw-2\fboxsep-2\fboxrule\relax][c]{\dimexpr\qualw-2\fboxsep-2\fboxrule\relax}{\centering\color{gray}#1}}}
\newcommand{\qualrow}[1]{\parbox[c][\qualw][c]{0.7cm}{\centering\rotatebox{90}{#1}}}
\begin{tabular}{@{}c@{\hspace{1.5pt}}c@{\hspace{1.5pt}}c@{\hspace{1.5pt}}c@{\hspace{1.5pt}}c@{\hspace{1.5pt}}c@{\hspace{1.5pt}}c@{}}
& \parbox[b]{\qualw}{\centering Context (last real frame)\strut\vspace{1pt}}
& \parbox[b]{\qualw}{\centering GT future ($+6$)\strut\vspace{1pt}}
& \parbox[b]{\qualw}{\centering Kernel + acc.\strut\vspace{1pt}}
& \parbox[b]{\qualw}{\centering W-Flow\strut\vspace{1pt}}
& \parbox[b]{\qualw}{\centering W-Flow + acc.\strut\vspace{1pt}}
& \parbox[b]{\qualw}{\centering\textbf{MSA (ours)}\strut\vspace{1pt}} \\
\qualrow{BridgeV2} & \qualimg{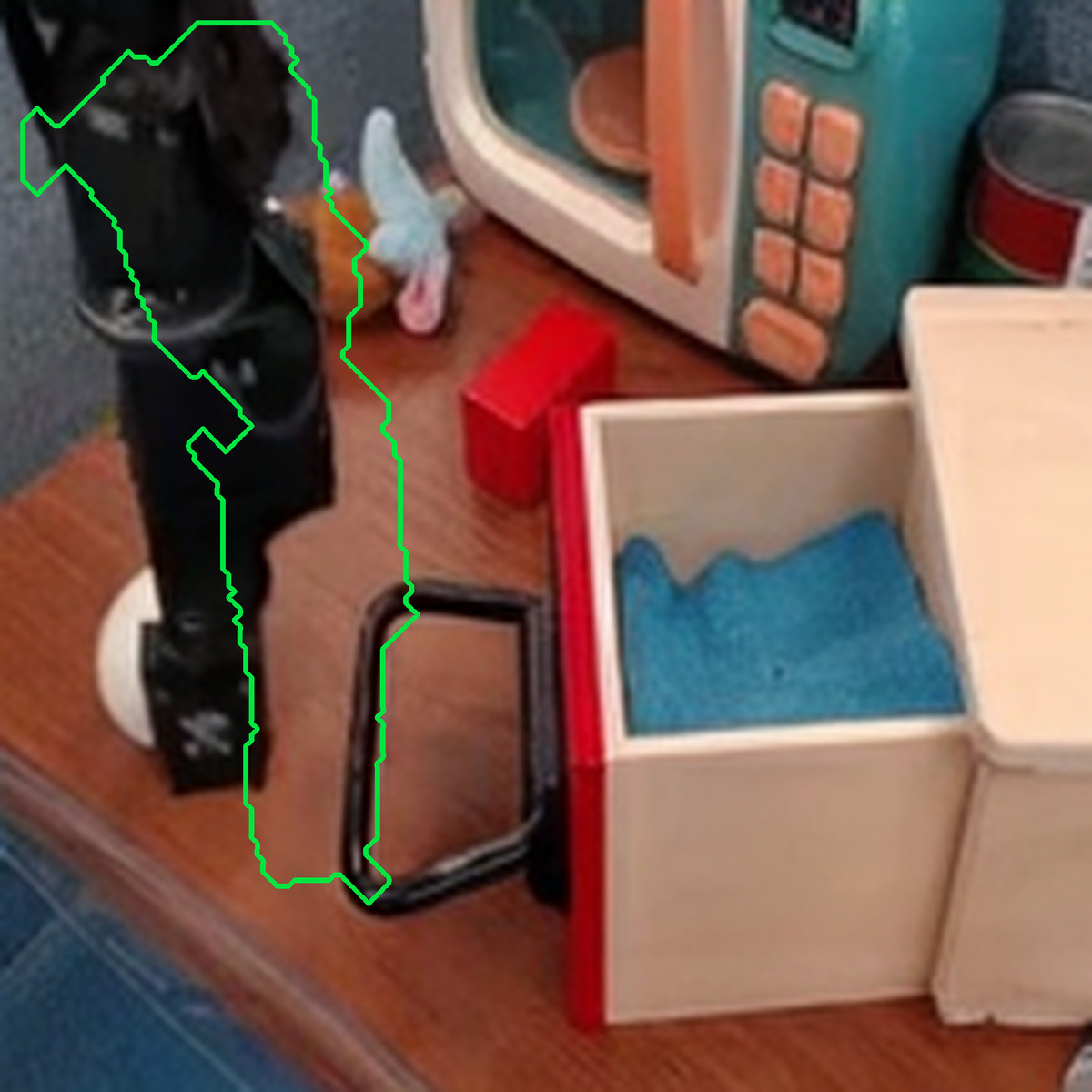} & \qualimg{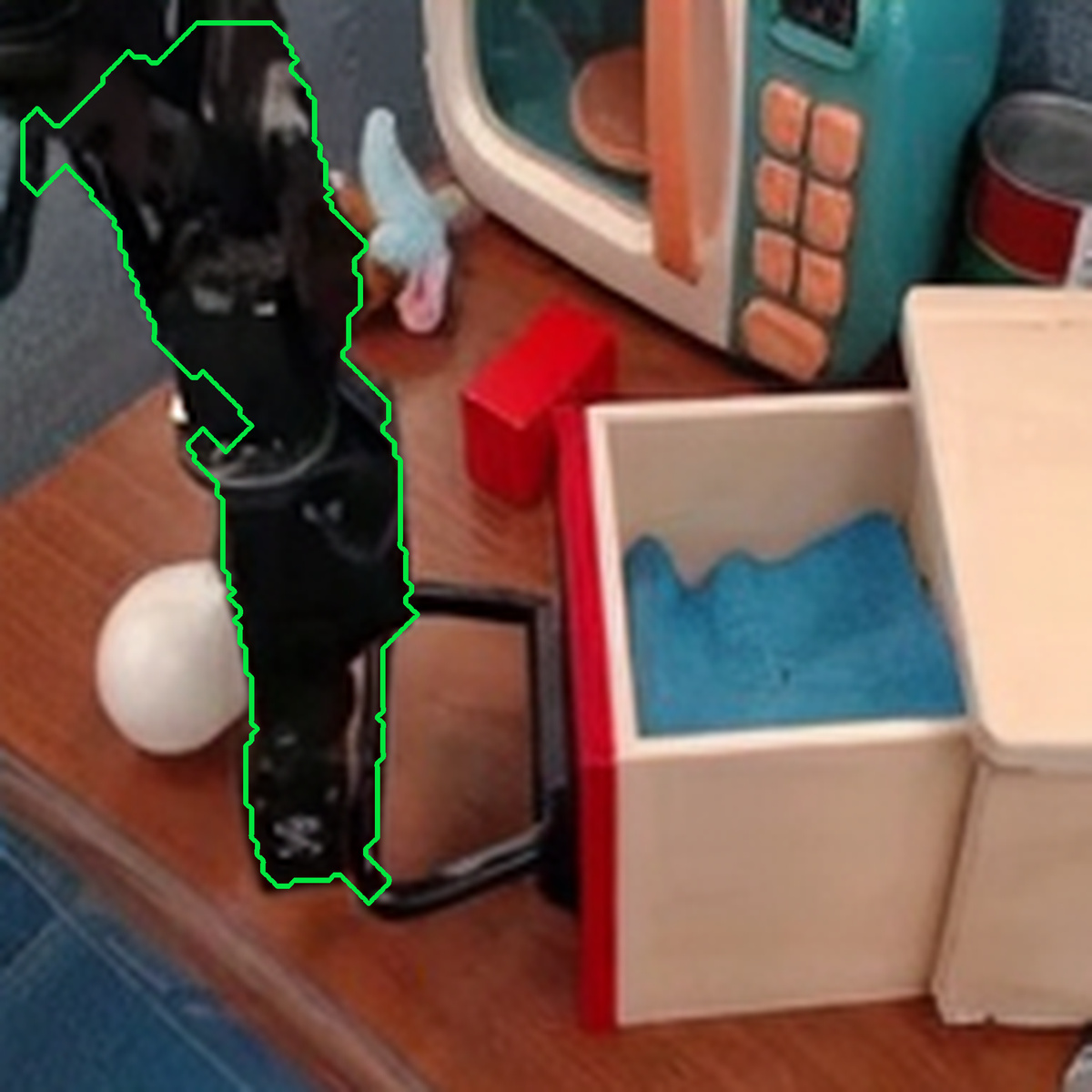} & \qualimg{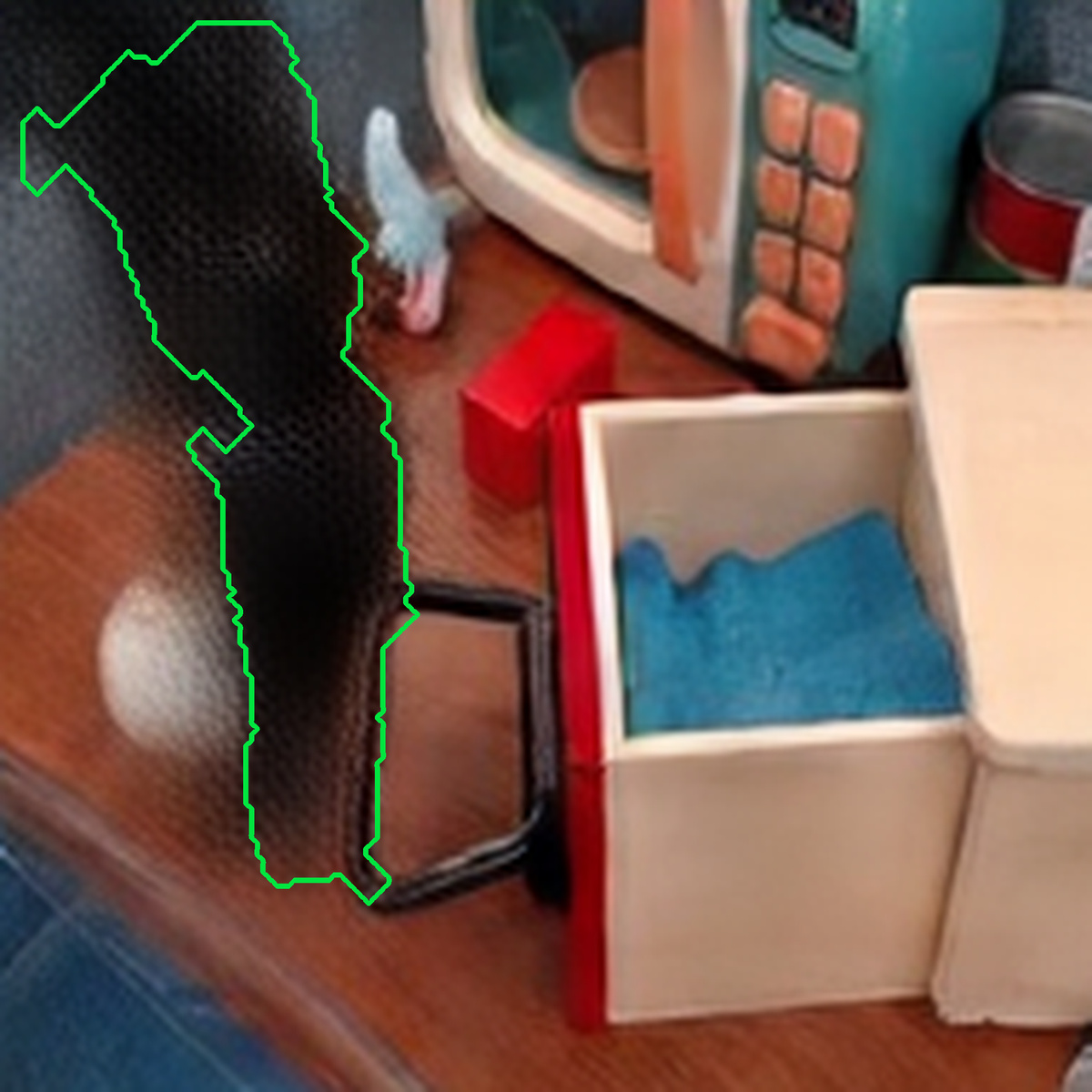} & \qualimg{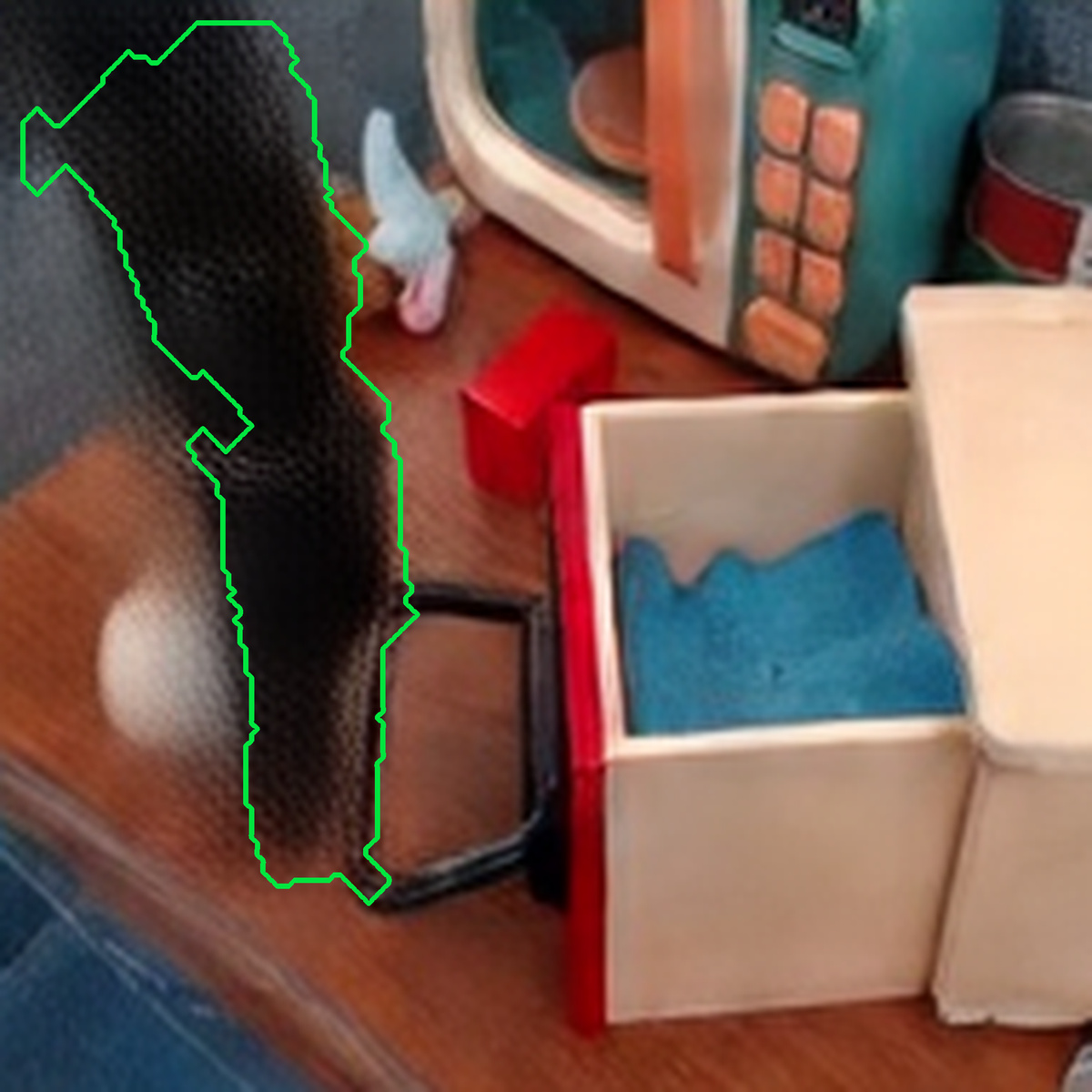} & \qualimg{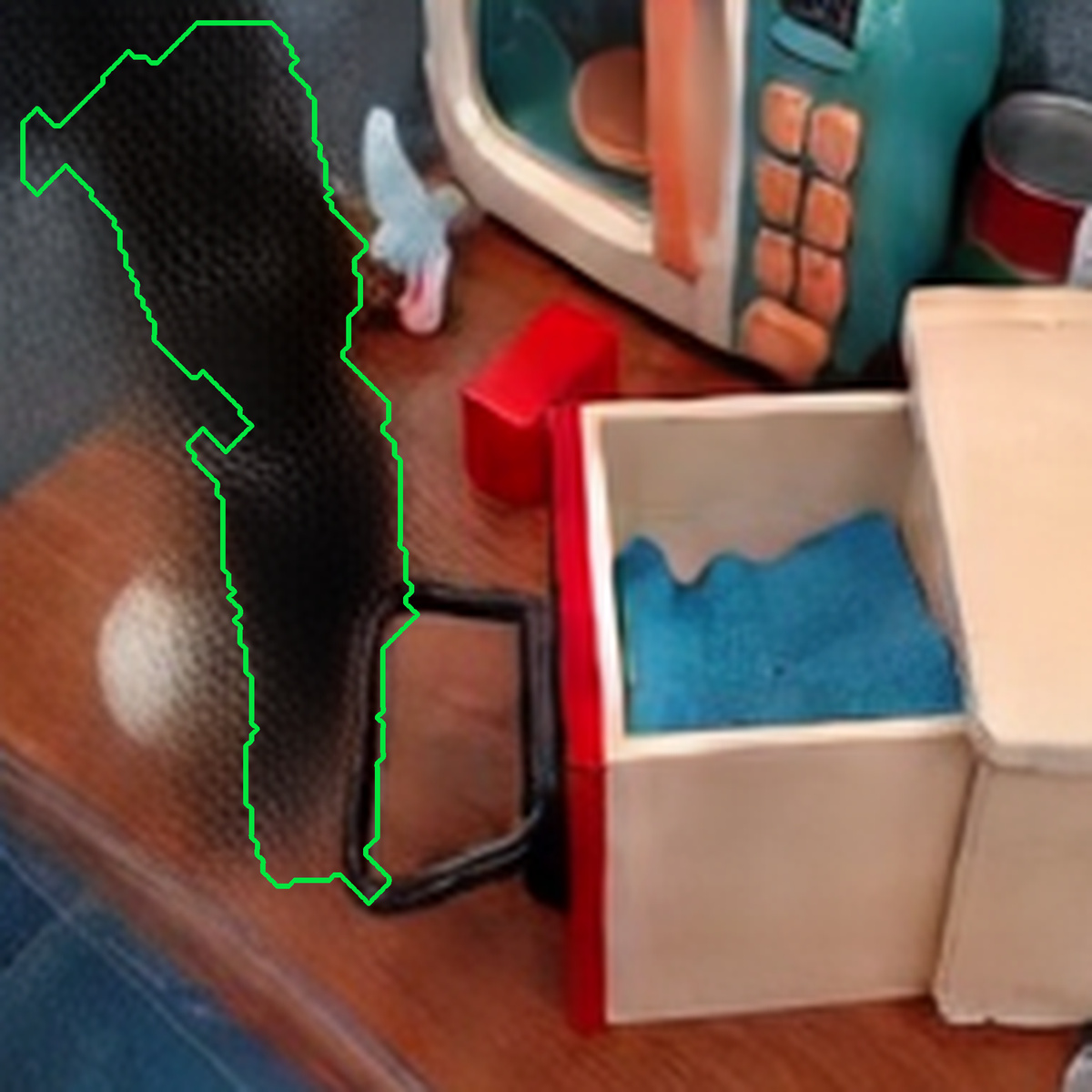} & \qualimg{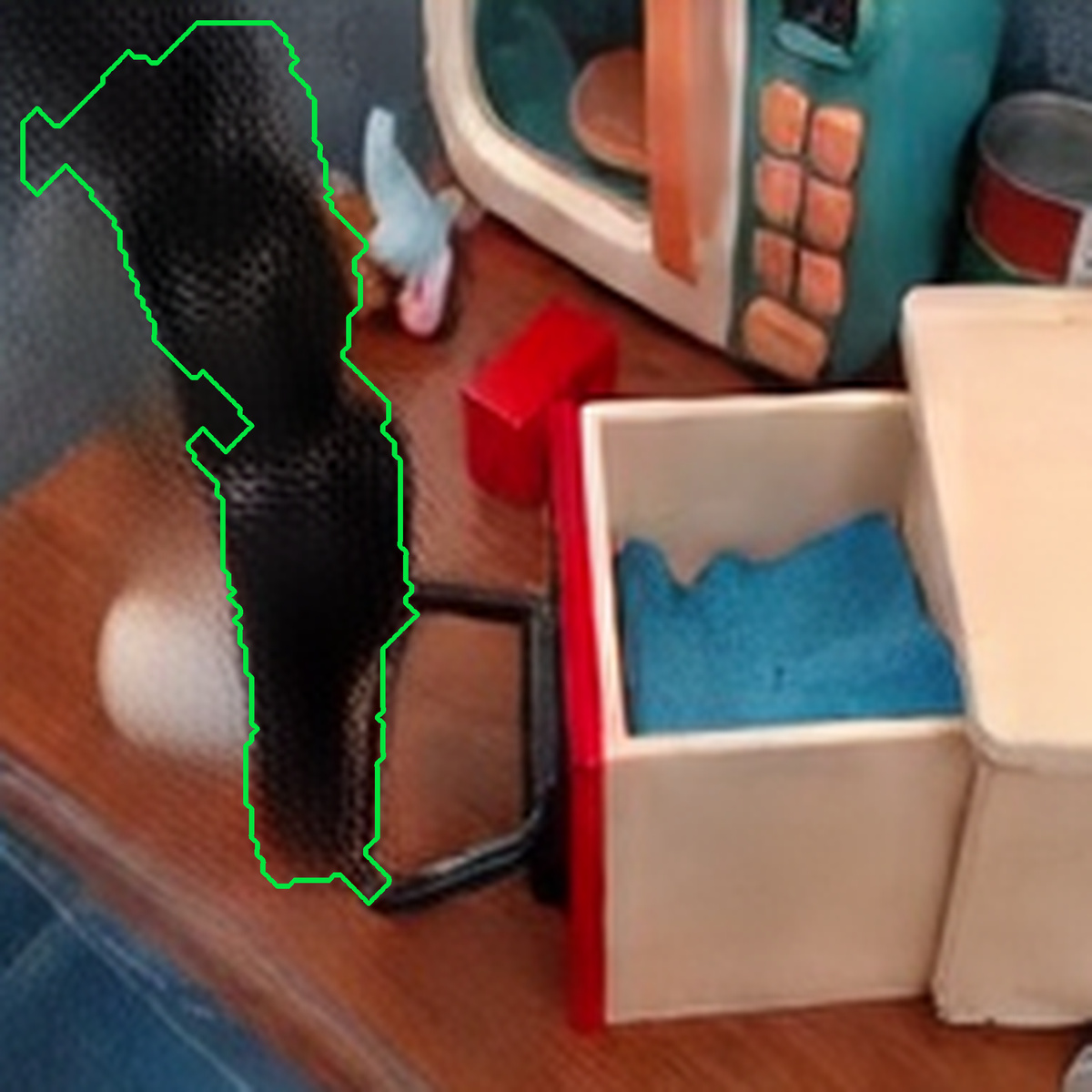} \\[1.5pt]
\qualrow{Push-T} & \qualimg{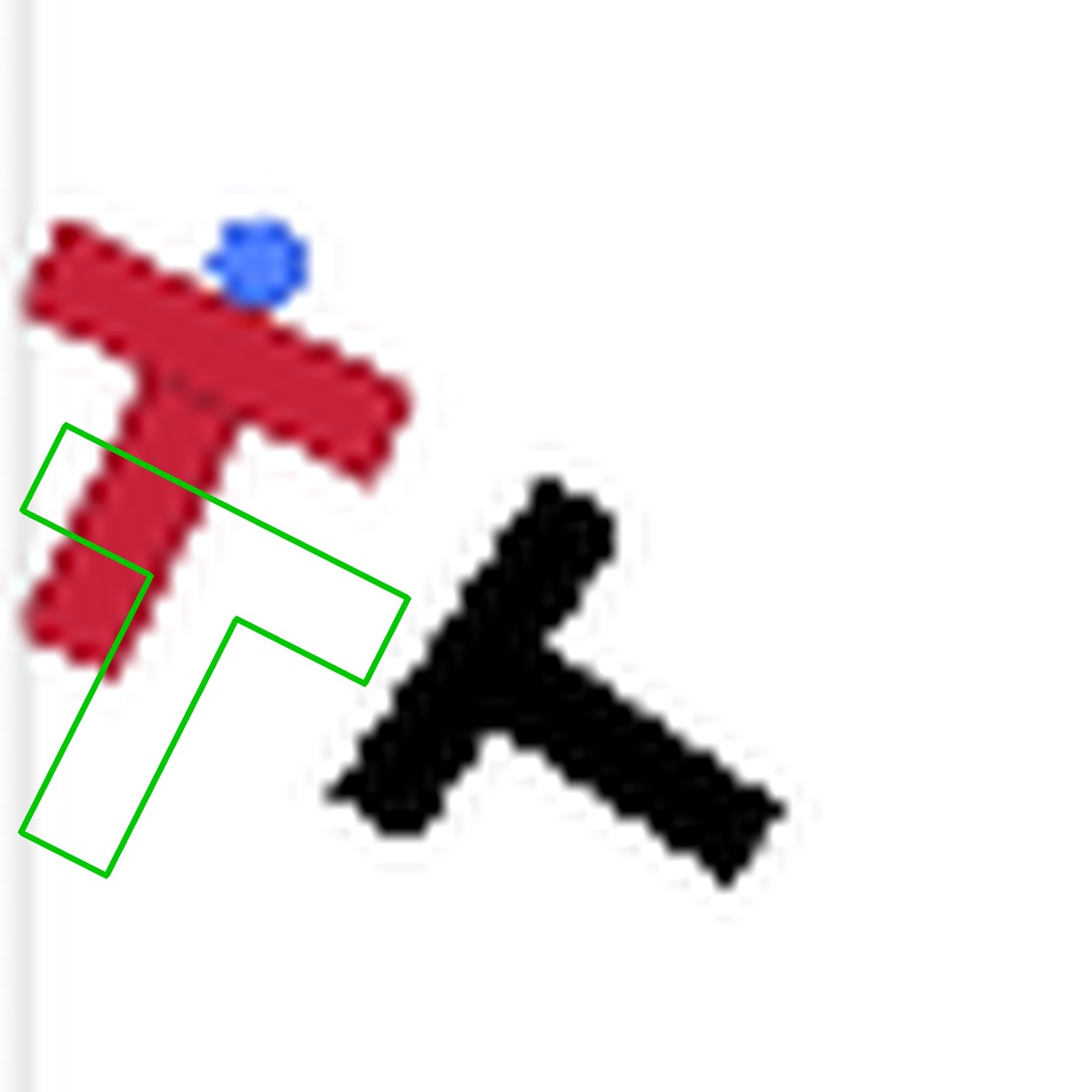} & \qualimg{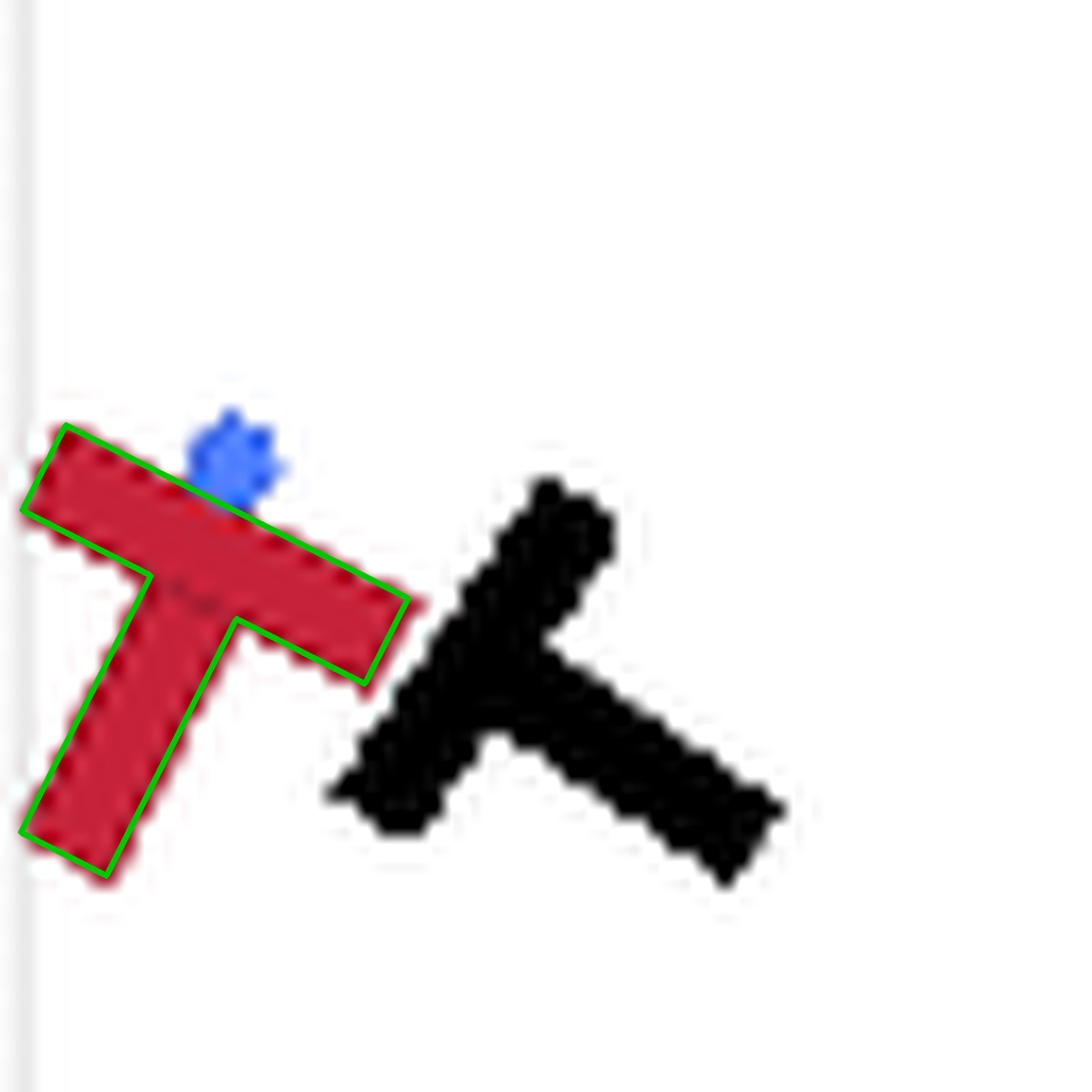} & \qualimg{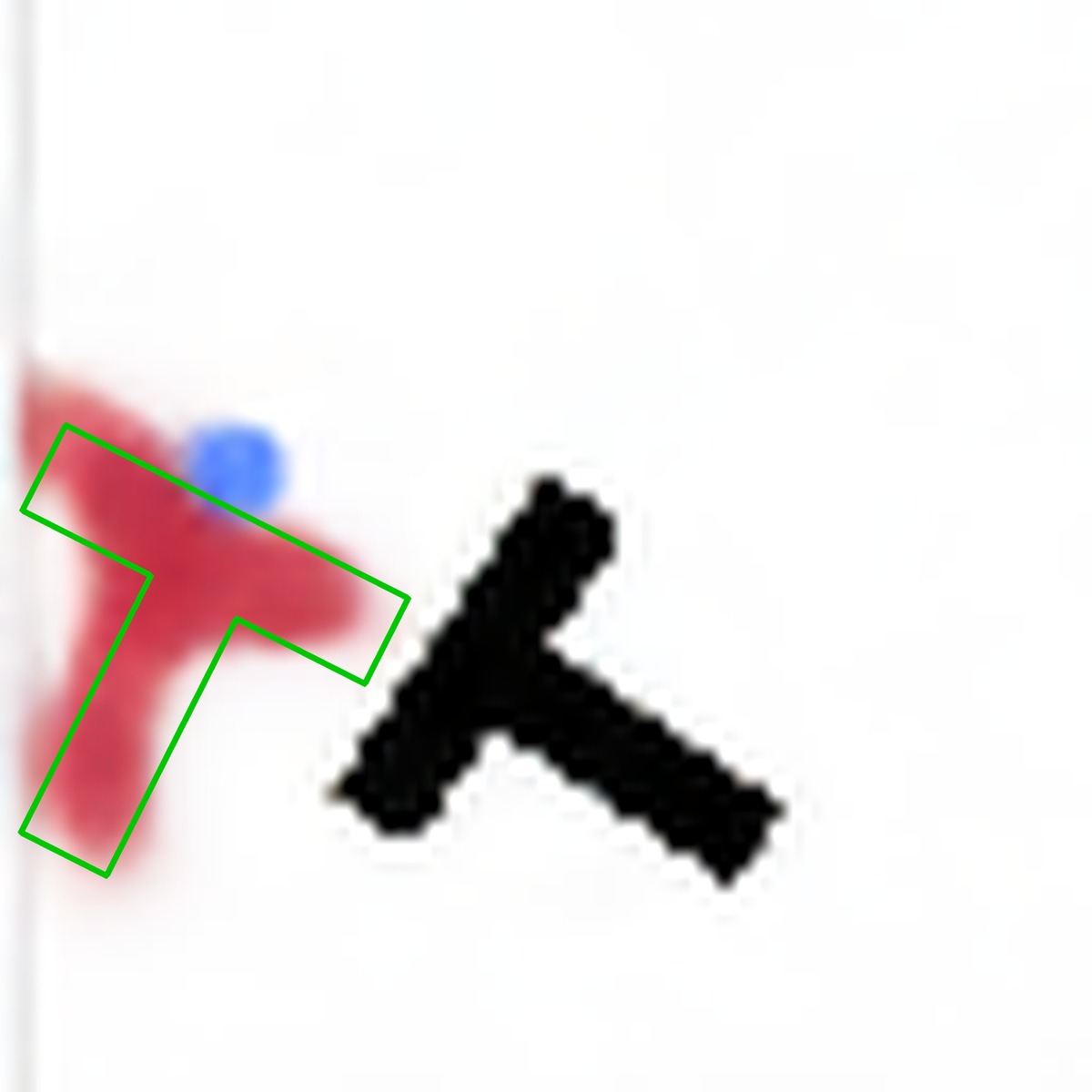} & \qualimg{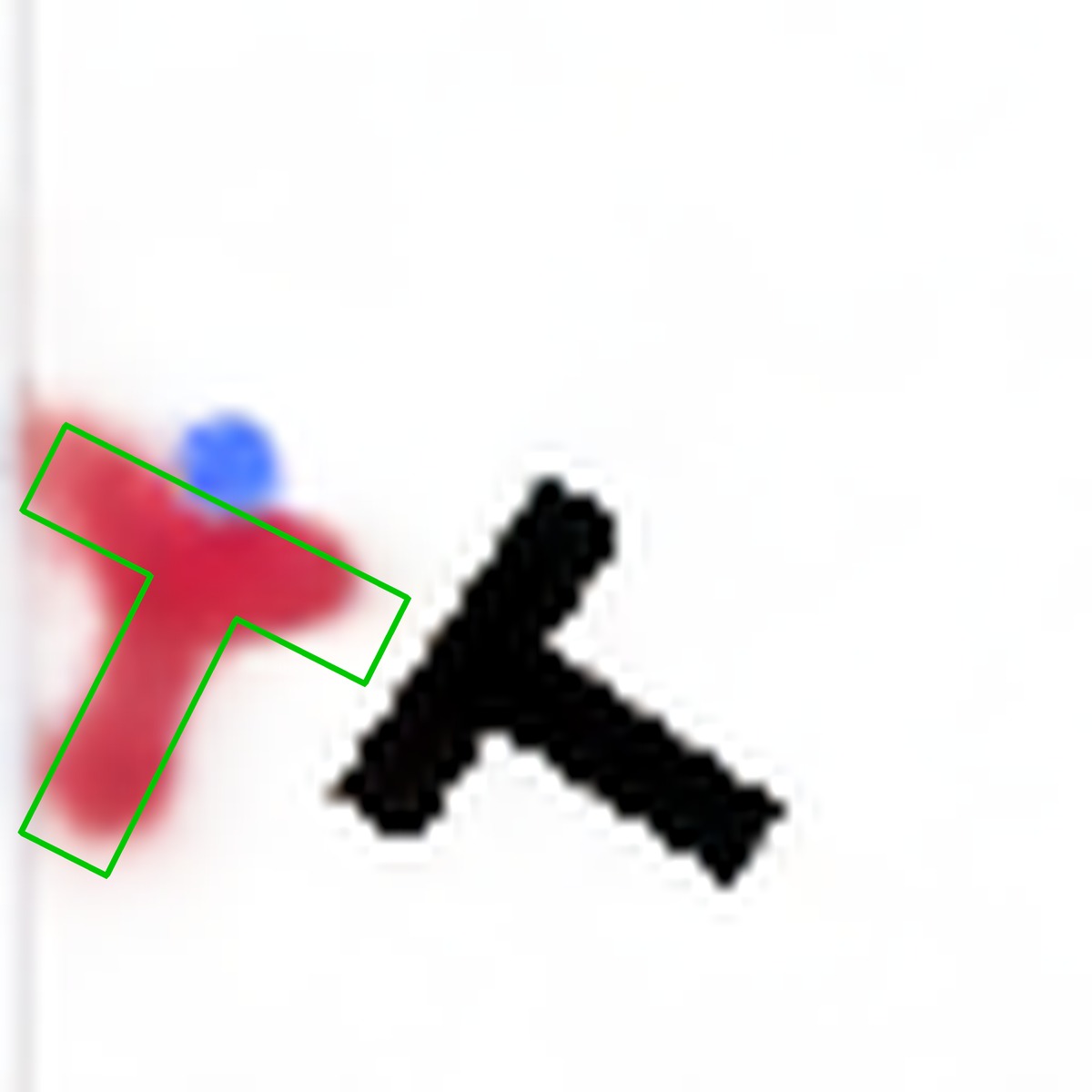} & \qualimg{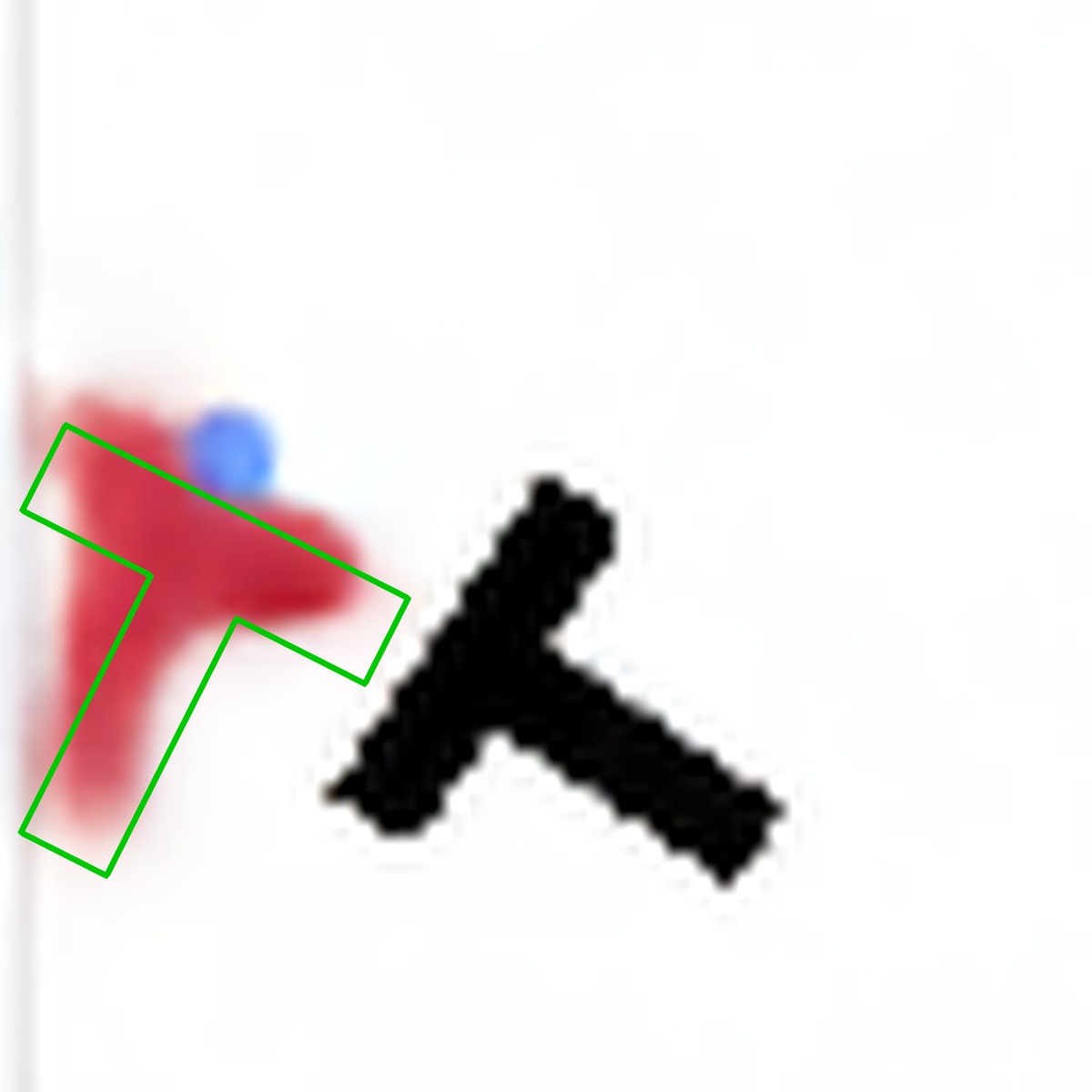} & \qualimg{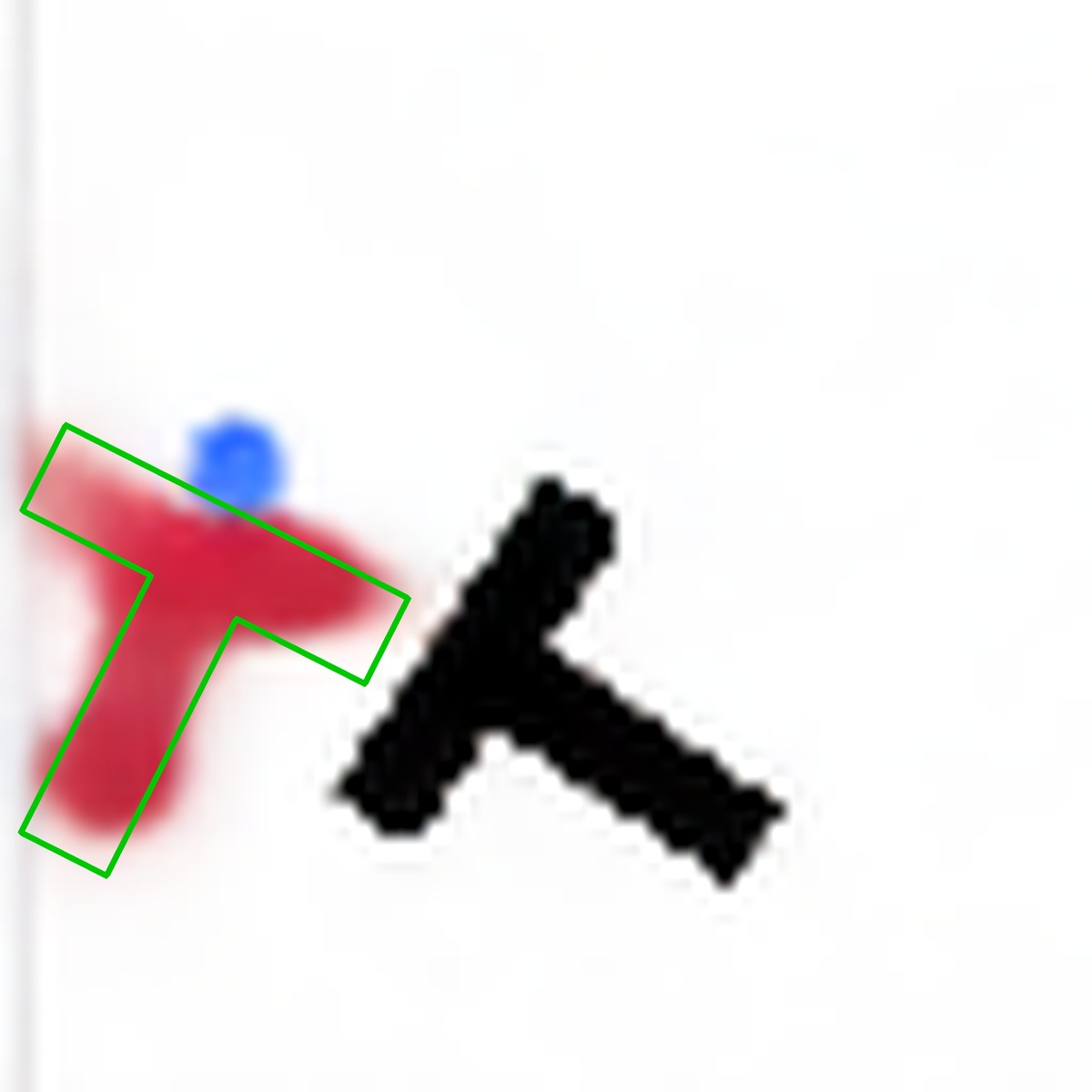} \\[1.5pt]
\qualrow{\shortstack{Robomimic\\Lift}} & \qualimg{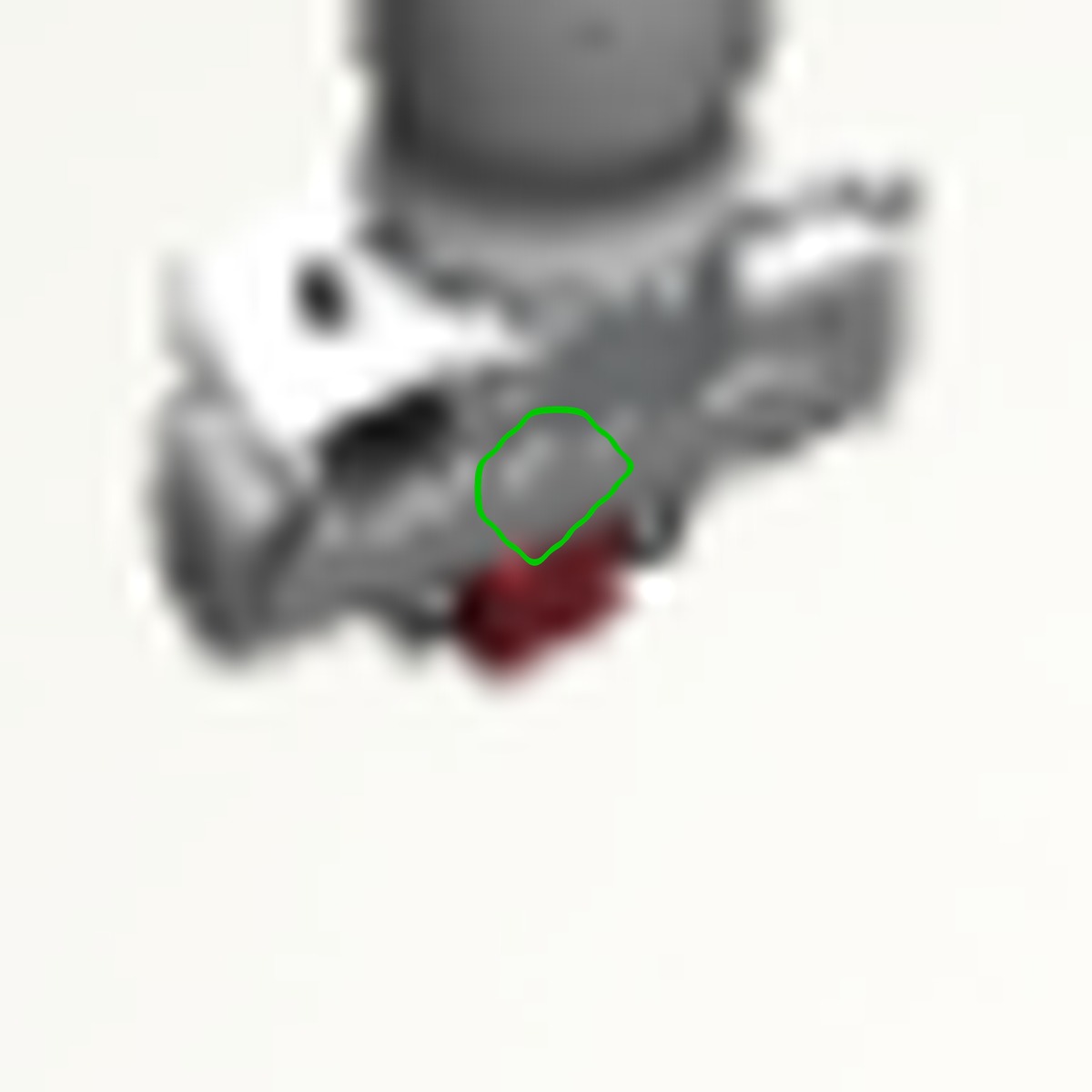} & \qualimg{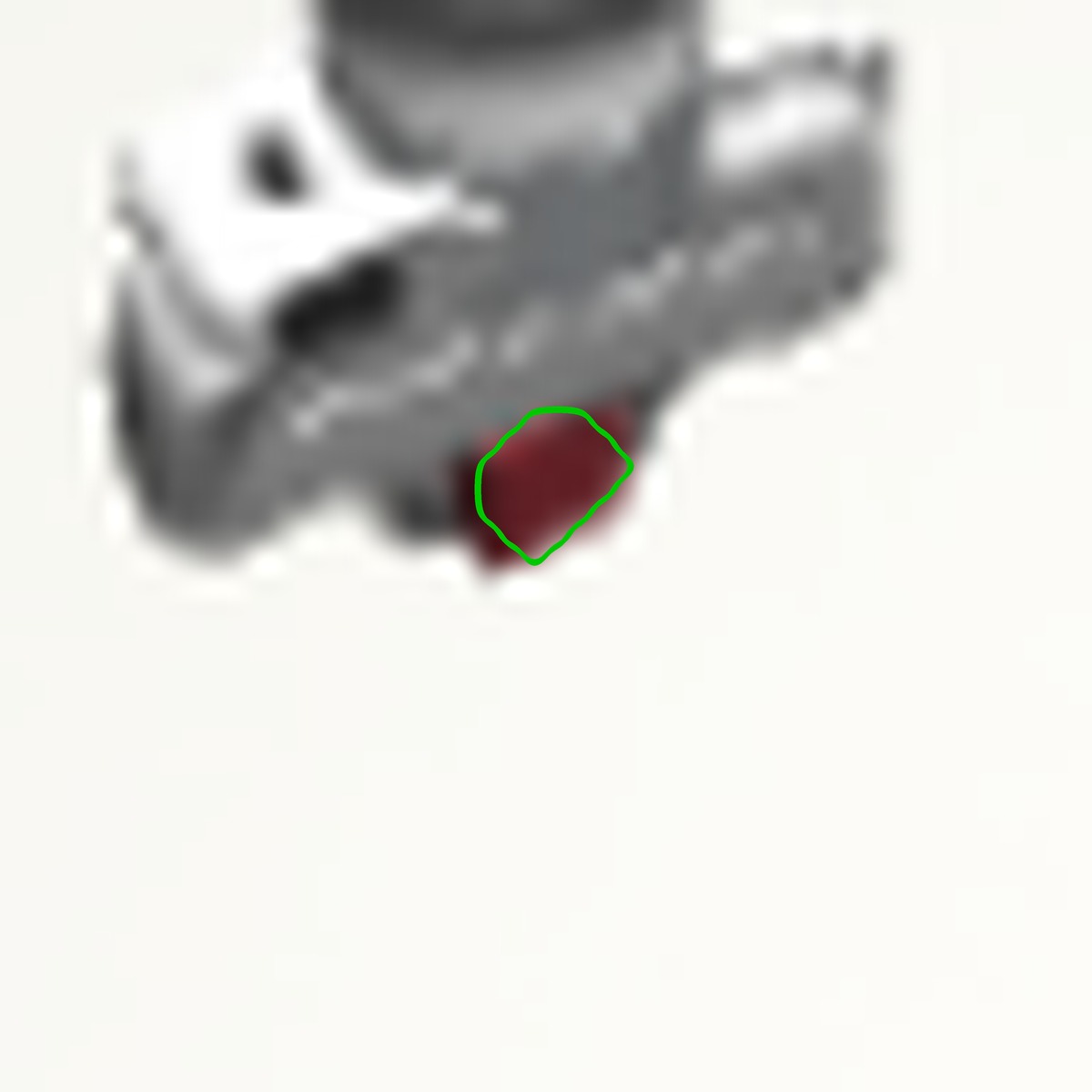} & \qualimg{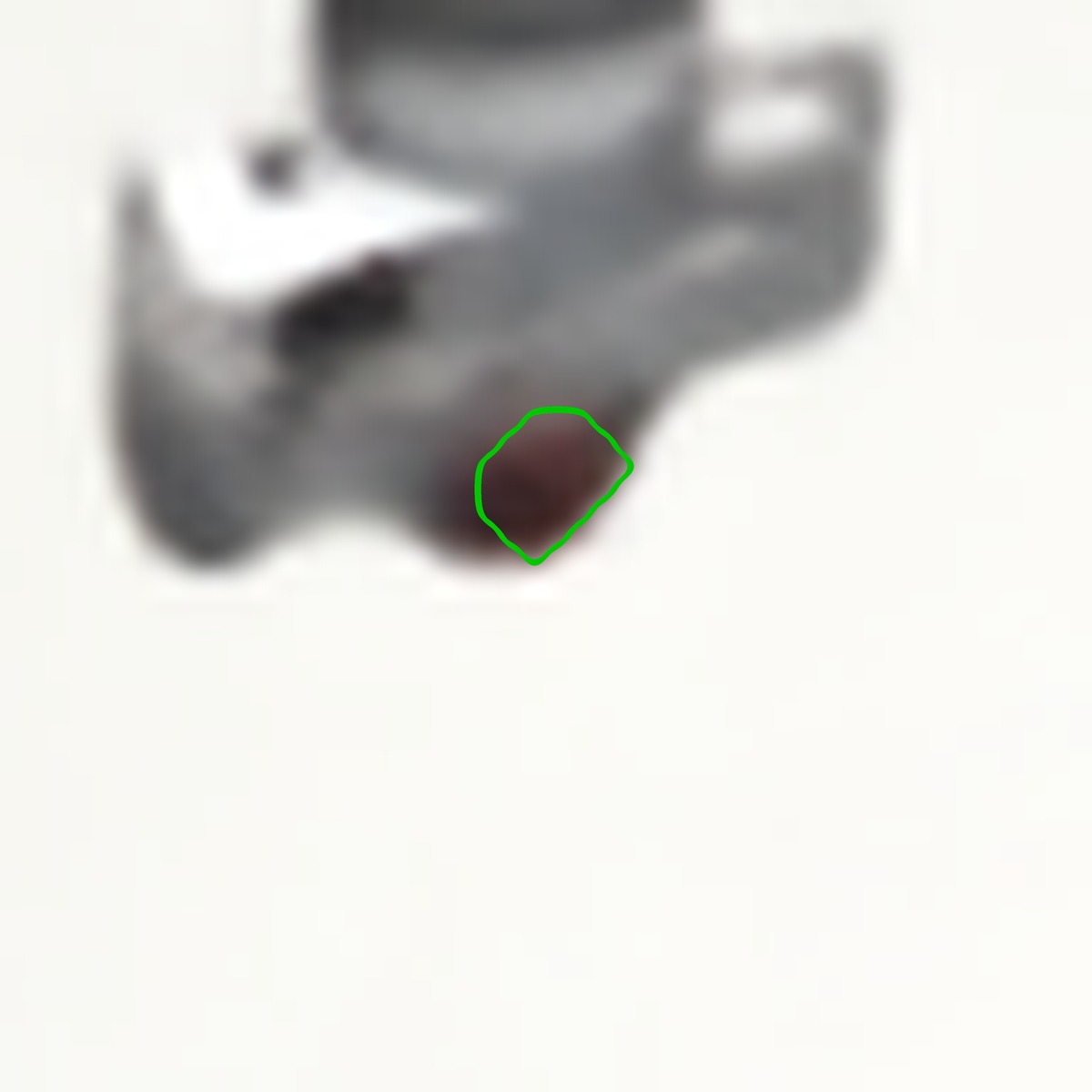} & \qualimg{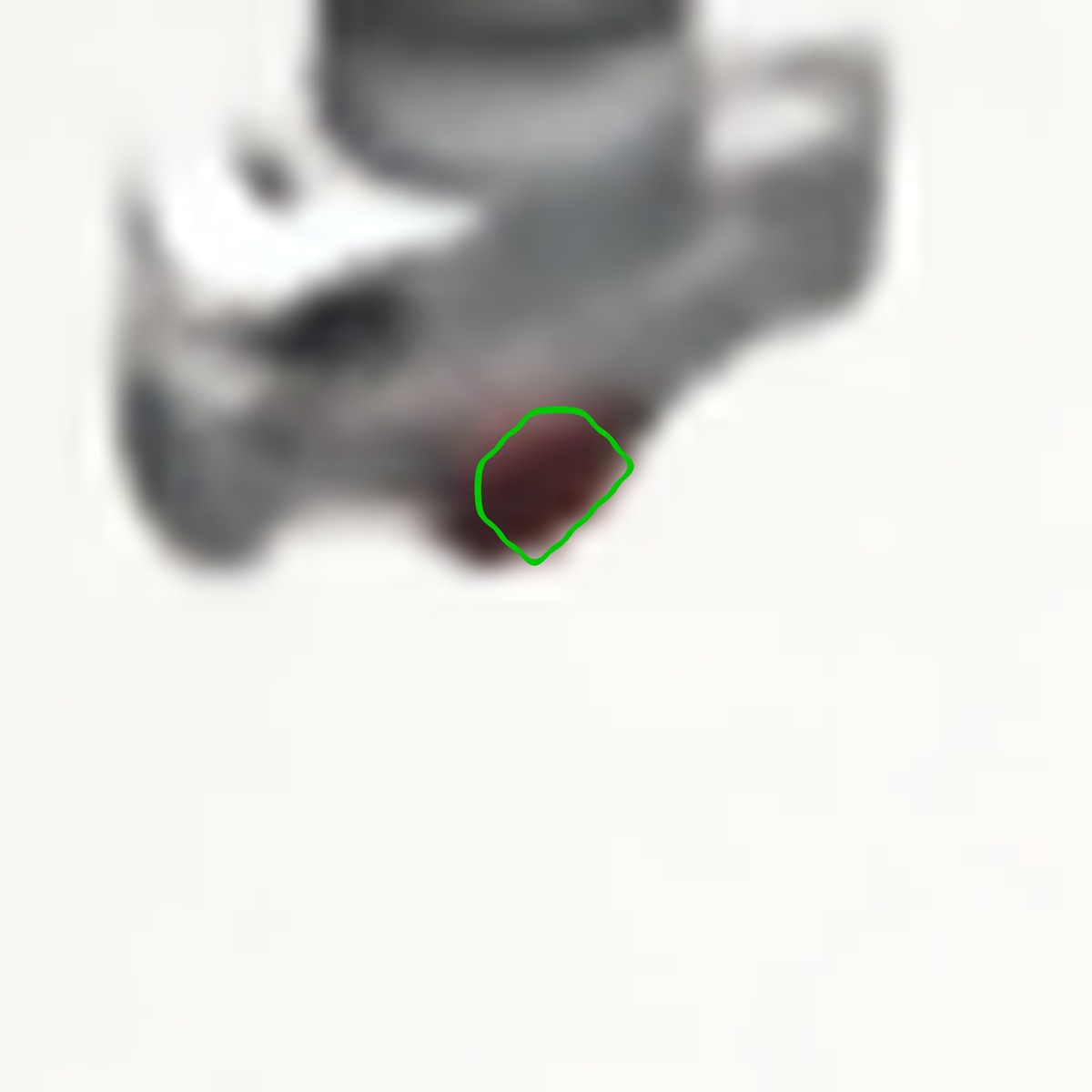} & \qualimg{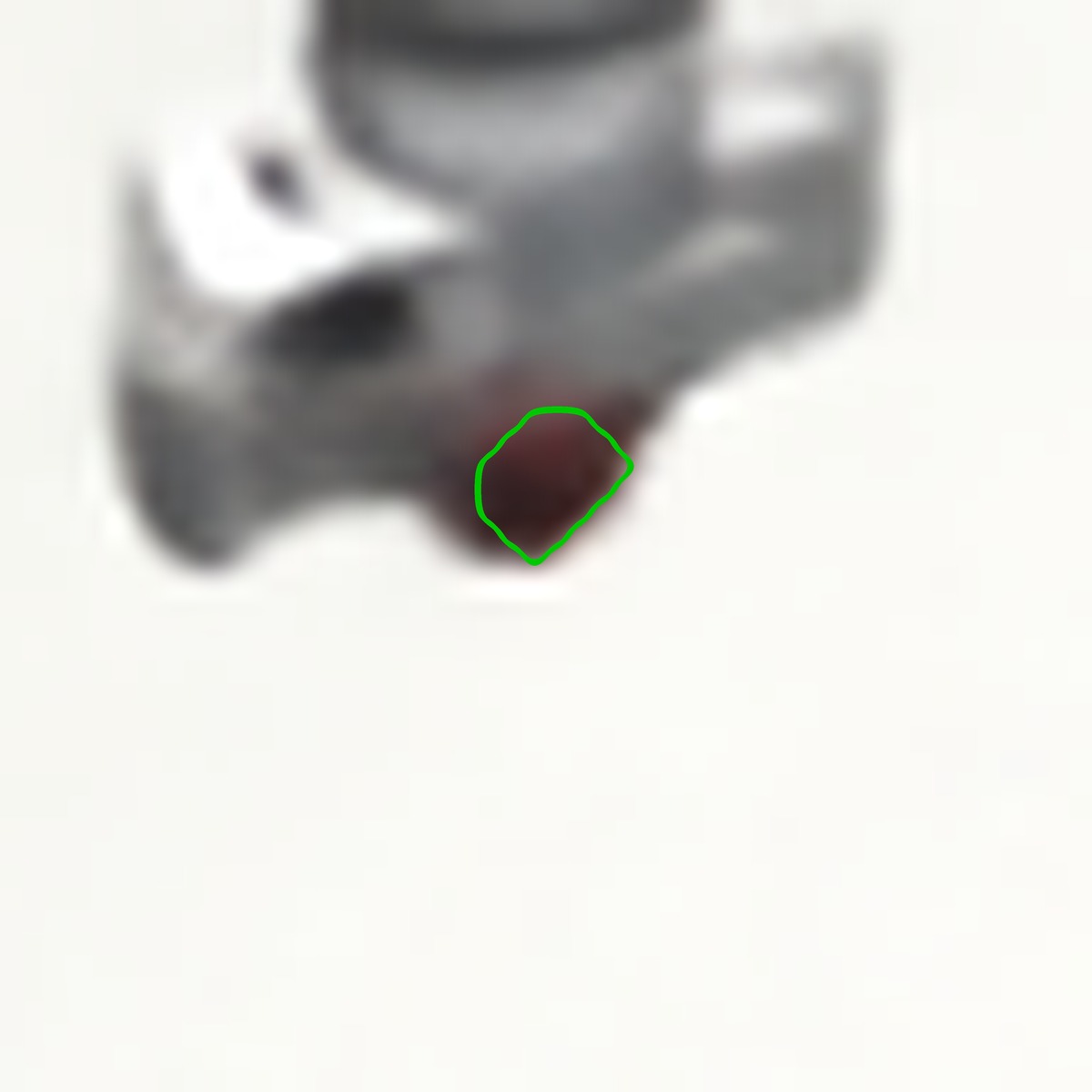} & \qualimg{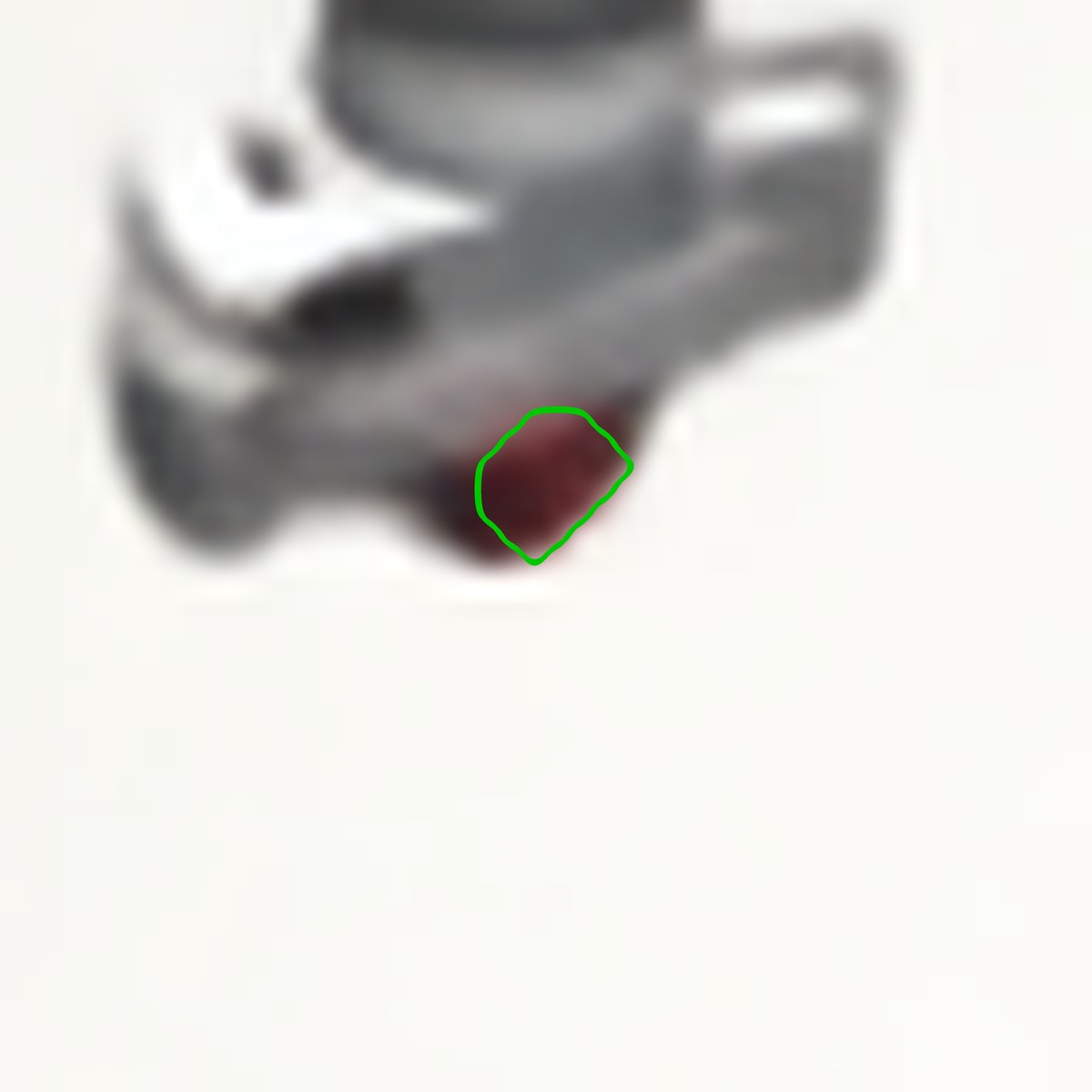} \\[1.5pt]
\qualrow{\shortstack{Robomimic\\Can}} & \qualimg{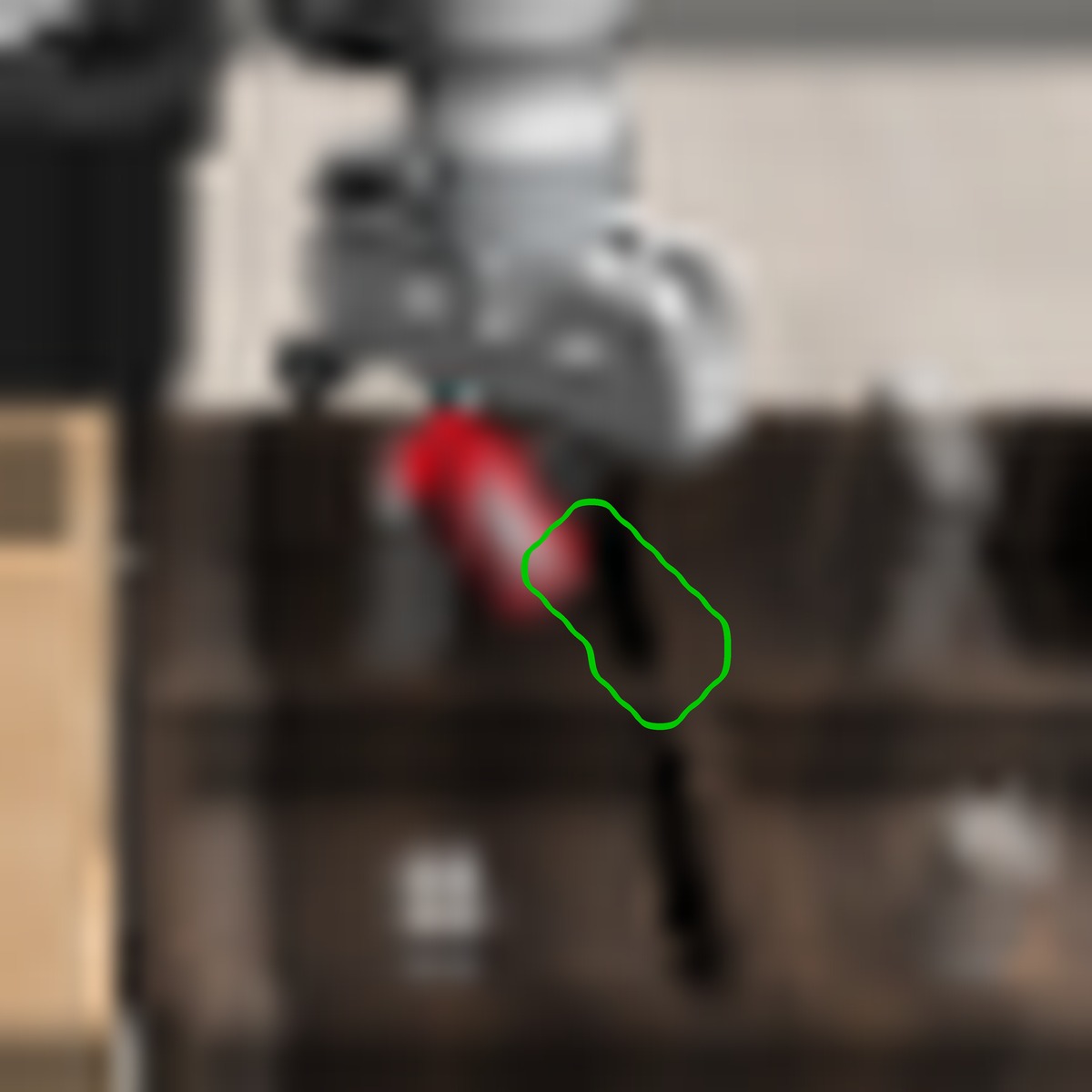} & \qualimg{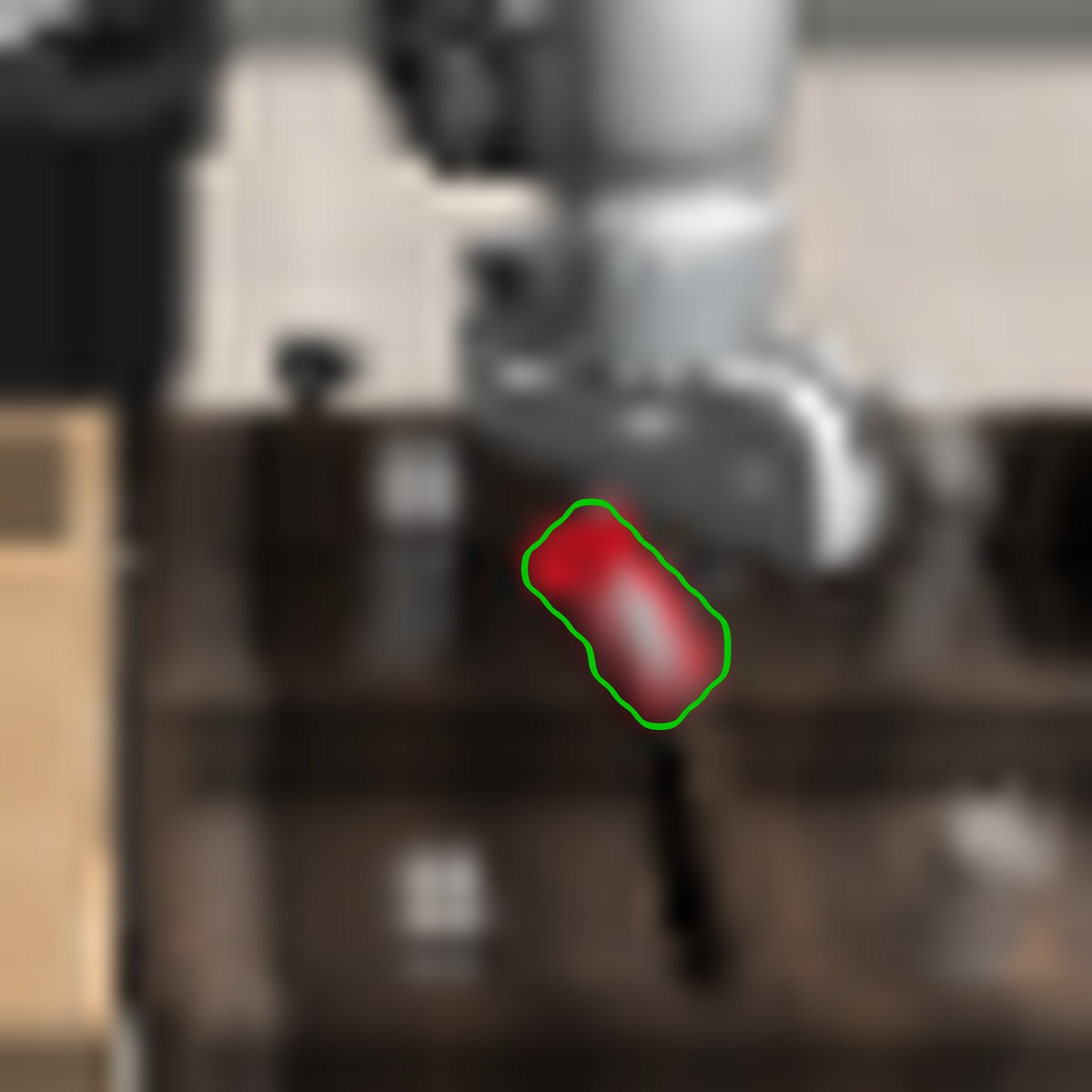} & \qualimg{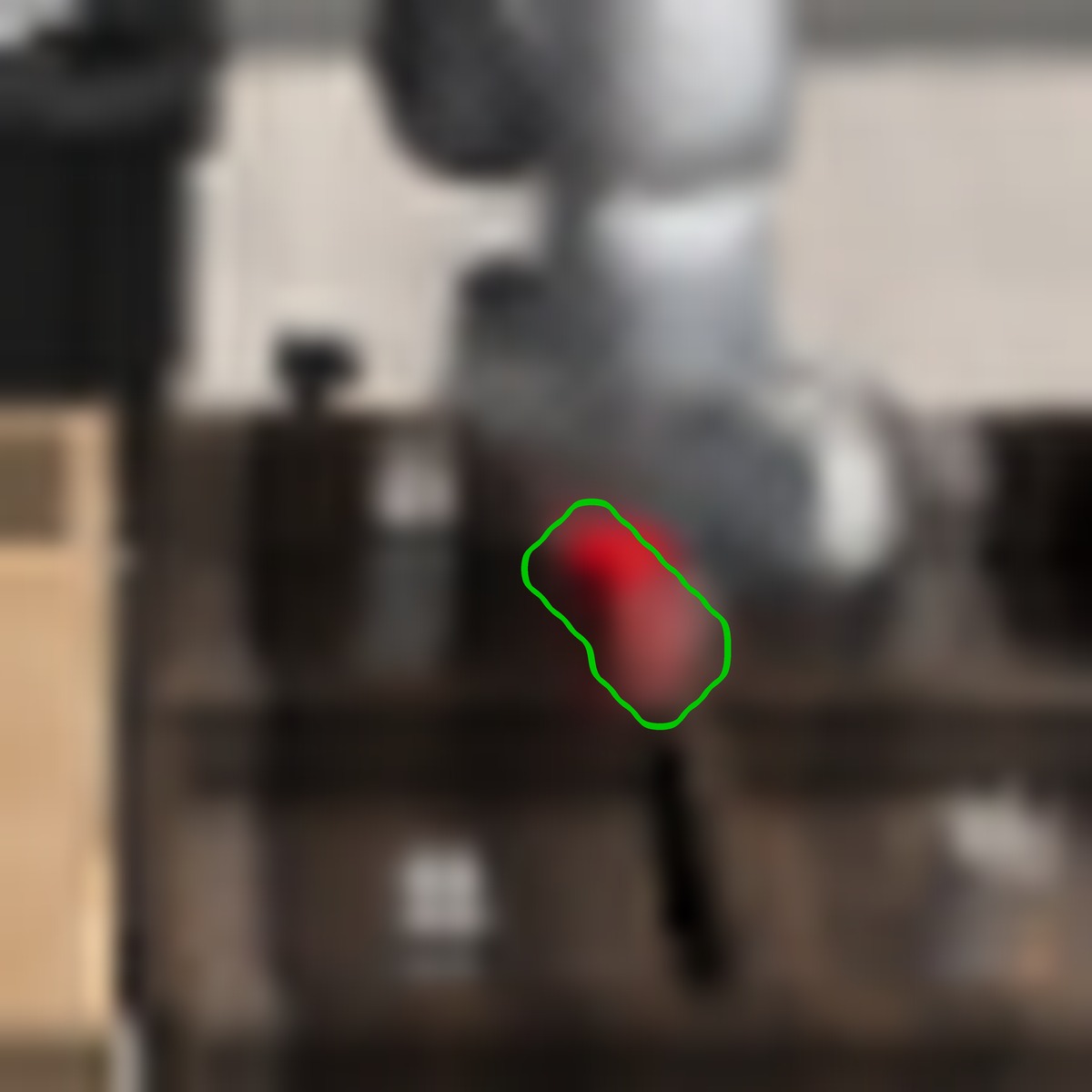} & \qualimg{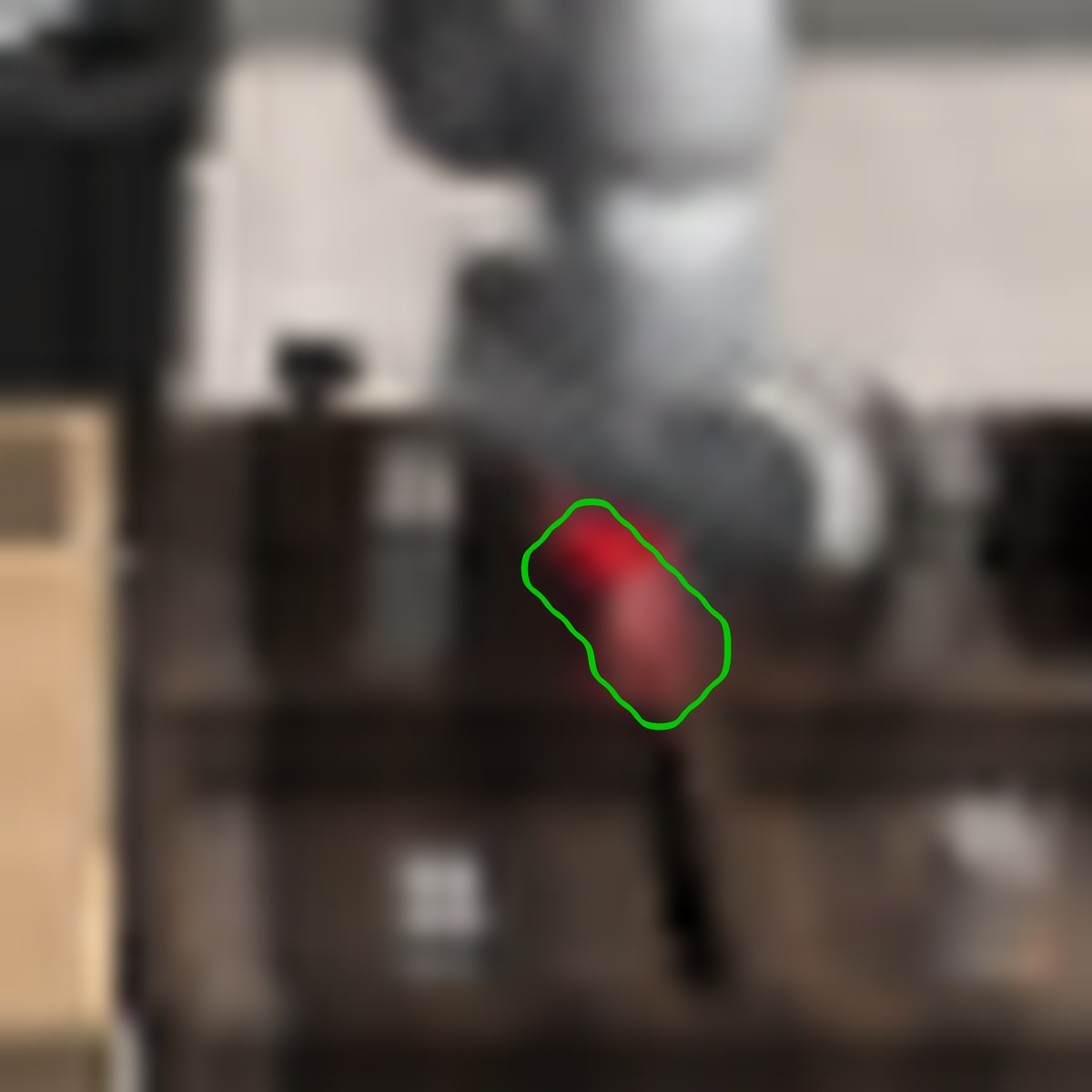} & \qualimg{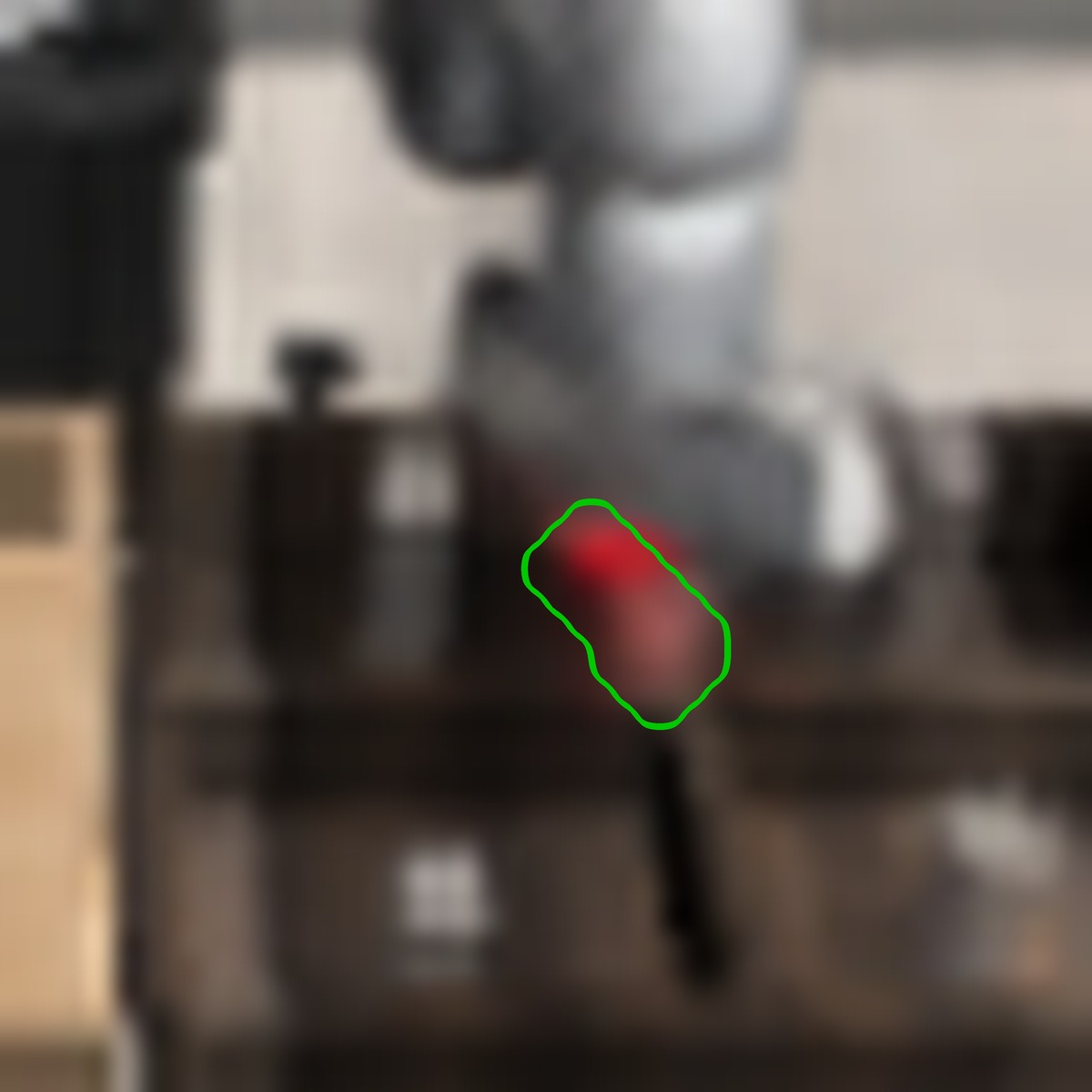} & \qualimg{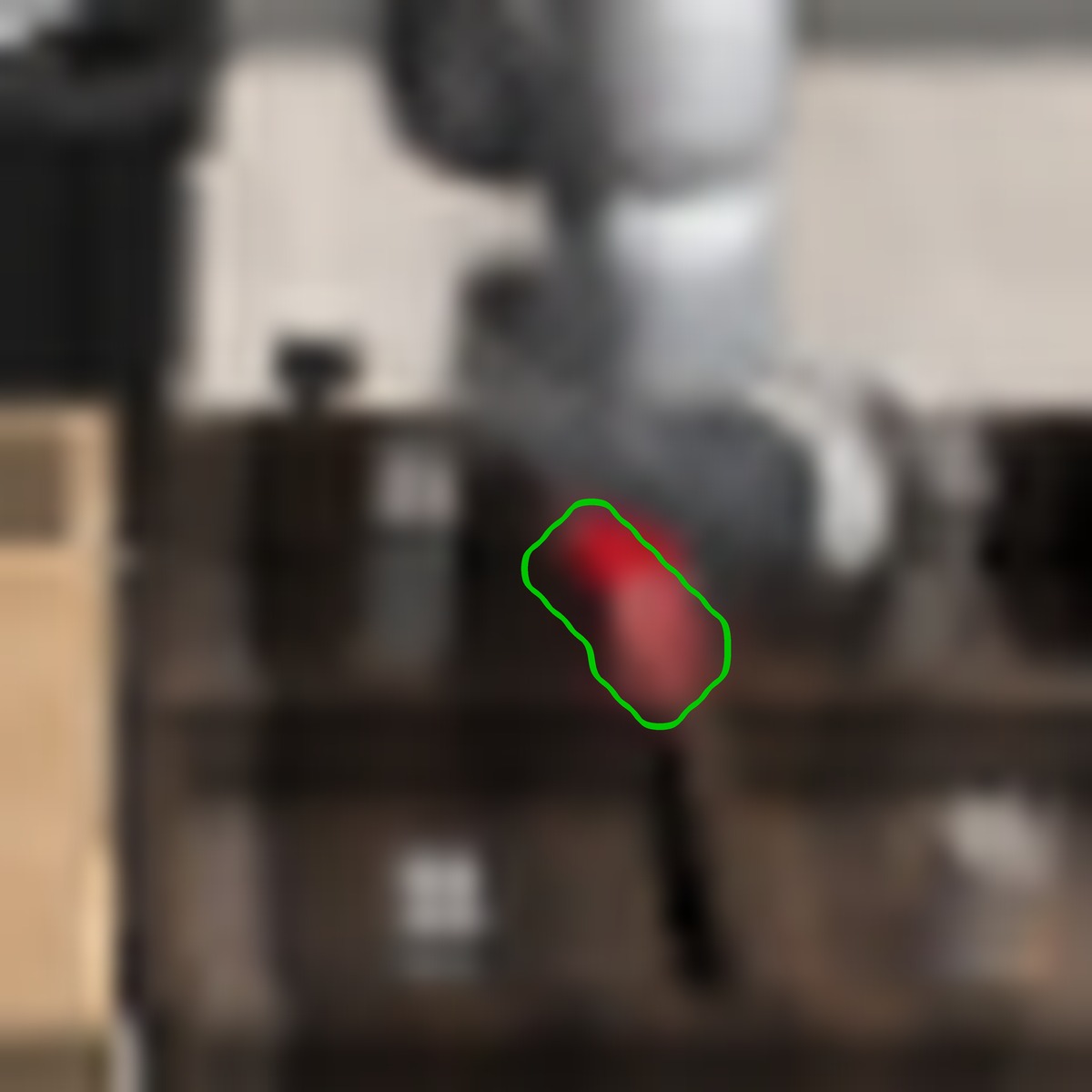} \\
\end{tabular}
\caption{\textbf{Qualitative rollouts.} Last context frame, ground truth six steps later, and each model's prediction under the recorded actions. BridgeV2: top-left $191\times191$ crop. The green contour is the ground-truth object (arm on BridgeV2) at $+6$.}
\label{fig:qual}
\end{figure*}

\begin{table*}[t]
\centering
\caption{\textbf{Results on BridgeV2.} Controllability: PSNR with true actions and its drop $\Delta$PSNR when shuffled (489 windows). Motion probe: progress along the real motion, off-motion error, and spread (200 windows $\times$ 16 samples). Quality: 8-frame rollouts on 491 validation episodes. Best among learned models in \textbf{bold}.}
\label{tab:bridge}
\footnotesize
\setlength{\tabcolsep}{3pt}
\begin{tabular}{l cc cccc ccccc}
\toprule
& \multicolumn{2}{c}{Controllability} & \multicolumn{4}{c}{Motion probe} & \multicolumn{5}{c}{Visual and temporal quality} \\
\cmidrule(lr){2-3}\cmidrule(lr){4-7}\cmidrule(lr){8-12}
Method & $\Delta$PSNR$\uparrow$ & PSNR$\uparrow$ & Prog.$\uparrow$ & Prog.$_{\text{mov}}\uparrow$ & Off-mot.$\downarrow$ & Spread & SSIM$\uparrow$ & PSNR$\uparrow$ & LPIPS$\downarrow$ & FID$\downarrow$ & FVD$\downarrow$ \\
\midrule
Copy-last-frame & 0 & 19.99 & 0 & 0 & -- & -- & .832 & 20.07 & .111 & 4.17 & 251.7 \\
Unrelated real video & -- & -- & -- & -- & -- & -- & .211 & 10.07 & .656 & 23.60 & 115.1 \\
\midrule
Kernel & 2.108 & 23.03 & 0.427 & 0.464 & 0.339 & 0.025 & .818 & 23.28 & .159 & 55.49 & 326.1 \\
Kernel + acc. & 2.111 & 23.00 & 0.436 & 0.450 & 0.347 & 0.025 & .831 & 23.23 & .148 & \textbf{55.44} & 304.4 \\
W-Flow & 2.185 & 22.99 & 0.426 & 0.440 & \textbf{0.326} & 0.031 & .827 & 23.18 & .148 & 56.40 & 323.3 \\
W-Flow + acc. & 2.029 & 22.81 & 0.443 & 0.456 & 0.398 & 0.026 & .825 & 23.02 & .148 & 56.30 & 332.9 \\
MSA (ours) & \textbf{2.401} & \textbf{23.24} & \textbf{0.533} & \textbf{0.549} & 0.337 & 0.026 & \textbf{.847} & \textbf{23.50} & \textbf{.142} & 58.83 & \textbf{243.1} \\
\bottomrule
\end{tabular}
\end{table*}

Each world model serves as the simulator of GPC-Rank~\cite{qi2026gpc} on Push-T, ranking 50 proposals of a diffusion policy~\cite{chi2023diffusionpolicy} (\textit{Supp.~B.8}).
In Tab.~\ref{tab:gpc}, MSA has the highest mean IoU for both policies (0.744 and 0.740 vs.\ at most 0.717 and 0.722).
With 50 shared episodes the standard error is 0.02 to 0.03 (s.d.$/\sqrt{50}$), so these gaps are about one standard error: planning agrees in direction with the $\tau$ gain without separating the models.

\subsection{Target bias and accentuation scale}
\label{sec:exp:alpha}

\paragraph{Controlled setting.}
Fig.~\ref{fig:hero} trains the same one-step generator with each field on a T-block world in which every model can reach the recorded future (\textit{Supp.~C}).
Tab.~\ref{tab:toy} confirms the theory to three decimals: balanced accentuation converges to $\bar t=\alpha$, the fixed point of Prop.~\ref{prop:bal_bias}, and kernel accentuation and MSA without debiasing carry the smaller biases of Props.~\ref{prop:dw_push} and~\ref{prop:bias_correction}(i), both growing with the scale.
MSA converges to the recorded future for every weight.

\paragraph{Push-T.}
Trained world models stop short of the target (all slopes are below 1), so the overshoot of the controlled setting does not appear; the sweep shows how each field trades response against accuracy, not target bias itself.
Fig.~\ref{fig:alpha} sweeps $\alpha$: kernel accentuation barely responds, and balanced accentuation reaches MSA's slope (0.746) only near $\alpha=1.75$, where its $\tau$ is about 0.67 and its moving-region MSE about 0.26, against 0.712 and 0.170 for MSA.
Up to $\alpha=3.25$ its outcome bias grows from 0.484 to about 0.62 and its moving-region MSE more than doubles.
No $\alpha$ of either baseline reaches MSA in $\tau$ or moving-region MSE, and MSA needs no sampling-time scale.

\begin{table}[t]
\centering
\caption{\textbf{Full-episode rollouts.} Same 1000 starts and recorded actions, about 250 frames; the block pose is read from the generated frames. Best in \textbf{bold}.}
\label{tab:fullep}
\footnotesize
\setlength{\tabcolsep}{2.5pt}
\begin{tabular}{l cccccc}
\toprule
Method & \begin{tabular}[b]{@{}c@{}}Agent\\err (px)$\downarrow$\end{tabular} & \begin{tabular}[b]{@{}c@{}}Block\\xy (px)$\downarrow$\end{tabular} & \begin{tabular}[b]{@{}c@{}}Block\\angle ($^\circ$)$\downarrow$\end{tabular} & \begin{tabular}[b]{@{}c@{}}EPE\\mov.$\downarrow$\end{tabular} & \begin{tabular}[b]{@{}c@{}}Mov.\\MSE$\downarrow$\end{tabular} & \begin{tabular}[b]{@{}c@{}}Static\\drift$\downarrow$\end{tabular} \\
\midrule
Kernel & 2.56 & 44.3 & 44.1 & 1.263 & .209 & .037 \\
Kernel + acc. & 2.20 & 44.7 & 44.6 & 1.238 & .207 & \textbf{.036} \\
W-Flow & 1.56 & 44.1 & 44.3 & 1.231 & .198 & \textbf{.036} \\
W-Flow + acc. & 1.54 & 45.4 & 44.1 & 1.268 & .206 & \textbf{.036} \\
MSA (ours) & \textbf{1.44} & \textbf{42.1} & \textbf{43.6} & \textbf{1.181} & \textbf{.182} & .037 \\
\bottomrule
\end{tabular}
\end{table}

\addtocounter{table}{1}
\begin{table*}[t]
\centering
\caption{\textbf{Comparison with UW-Flow.} Cost and properties, with metrics at 15k and 40k steps (noise in px). UW-Flow + acc.\ matches $\lambda$ across three transport problems; MSA adds a closed-form gate. Best in \textbf{bold}.}
\label{tab:uwflow}
\footnotesize
\setlength{\tabcolsep}{3.5pt}
\begin{tabular}{l l c c @{\hspace{12pt}}ccccc@{\hspace{12pt}}cccc}
\toprule
& & & & \multicolumn{5}{c}{15k} & \multicolumn{4}{c}{40k} \\
\cmidrule(lr){5-9}\cmidrule(lr){10-13}
Method & Extra hyperparameters & \begin{tabular}[b]{@{}c@{}}Sinkhorn\\solves/step\end{tabular} & \begin{tabular}[b]{@{}c@{}}Zero push\\at target\end{tabular} & $\tau\uparrow$ & Slope$\uparrow$ & Noise$\downarrow$ & \begin{tabular}[b]{@{}c@{}}Mov.\\MSE$\downarrow$\end{tabular} & \begin{tabular}[b]{@{}c@{}}EPE\\mov.$\downarrow$\end{tabular} & $\tau\uparrow$ & Slope$\uparrow$ & Noise$\downarrow$ & \begin{tabular}[b]{@{}c@{}}Tail\\share$\downarrow$\end{tabular} \\
\midrule
W-Flow & -- & 2 & n/a & .301 & .349 & 3.30 & .303 & 1.695 & .538 & .579 & 2.33 & \textbf{.000} \\
W-Flow + acc. & $\alpha$ (train + inference) & 3 & $\times$ & .266 & .279 & \textbf{2.74} & .312 & 1.696 & .471 & .446 & \textbf{1.81} & \textbf{.000} \\
UW-Flow & $\lambda$ & 2 & n/a & .266 & .341 & 7.80 & .314 & 1.724 & .471 & .546 & 5.15 & \textbf{.000} \\
UW-Flow + acc. & $\lambda$ (matched $\times3$), $\alpha$ & 3 & $\times$ & \textbf{.320} & .337 & 3.67 & \textbf{.286} & 1.659 & .177 & .364 & 39.8 & .542 \\
MSA (ours) & $\beta$ fixed (no knob) & 2 + gate & $\checkmark$ & .315 & \textbf{.397} & 3.63 & \textbf{.286} & \textbf{1.610} & \textbf{.553} & \textbf{.612} & 2.23 & \textbf{.000} \\
\bottomrule
\end{tabular}
\end{table*}

\addtocounter{table}{-2}
\begin{table}[t]
\centering
\caption{\textbf{Gate components.} Accentuation added to W-Flow as $(\alpha-1)\,d$, $\beta\,r_i\,d$, and $\beta\,G\,d$ (\cref{eq:msa_guidance}). P/Z/S: progress-aware, zero at target, shape-neutral (\cref{sec:method:bias}). Best in \textbf{bold}.}
\label{tab:gate}
\footnotesize
\setlength{\tabcolsep}{2.1pt}
\begin{tabular}{l c ccccc}
\toprule
Variant & P\,/\,Z\,/\,S & $\tau\uparrow$ & Slope$\uparrow$ & Noise$\downarrow$ & Bias$\downarrow$ & Mov.\ MSE$\downarrow$ \\
\midrule
W-Flow & --\,/\,--\,/\,-- & 0.693 & 0.713 & 1.49 & -- & 0.1720 \\
+ constant push & $\times$\,/\,$\times$\,/\,$\checkmark$ & 0.678 & 0.698 & \textbf{1.32} & 0.484 & 0.1761 \\
+ joint softmax & $\checkmark$\,/\,$\times$\,/\,$\times$ & 0.694 & 0.718 & 1.44 & 0.492 & 0.1743 \\
MSA (ours) & $\checkmark$\,/\,$\checkmark$\,/\,$\checkmark$ & \textbf{0.712} & \textbf{0.746} & 1.55 & \textbf{0.462} & \textbf{0.1699} \\
\bottomrule
\end{tabular}
\end{table}
\addtocounter{table}{1}

\subsection{Full-episode rollouts}
\label{sec:exp:fullep}

Table~\ref{tab:fullep} rolls out full Push-T episodes (about 250 frames, no re-anchoring; \textit{Supp.~B.7}).
Block position errors (42.1 to 45.4~px) are large and do not separate the models; MSA has the lowest agent error (1.44 vs.\ 1.54~px) and moving-region EPE and MSE (.182 vs.\ .198), with static drift within .001.

\subsection{Gate components}
\label{sec:exp:gate}

In Tab.~\ref{tab:gate}, the constant push (W-Flow + acc.\ of Tab.~\ref{tab:main}) is neither progress-aware (D2) nor zero at the target (D1), and lowers the $\tau$ of W-Flow (0.693 to 0.678) and its slope.
The joint softmax is MSA without floor subtraction and sharing (same fixed weight, no $\alpha$ input): each prediction is pushed by its own share, $\beta\,r_i\,d$.
It edges W-Flow in $\tau$ and slope (by 0.001 and 0.005) but raises its moving-region MSE and has the highest outcome bias (0.492).
Subtracting the floor $r^+$ and sharing the push, the only difference from MSA, add 0.018 in $\tau$ and 0.028 in slope and give the lowest bias (0.462) and moving-region MSE, at the highest noise (1.55 vs.\ 1.32 to 1.49~px); \textit{Supp.~E.3} ablates the two separately.

\subsection{Gate annealing and response calibration}
\label{sec:exp:anneal}

\begin{figure}[t]
\centering
\includegraphics[width=\linewidth]{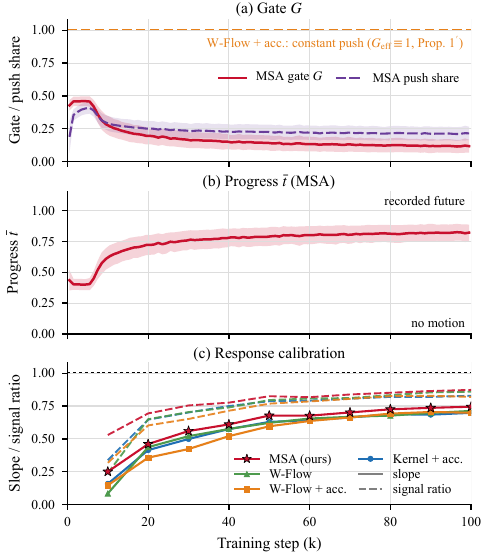}
\caption{\textbf{Gate annealing.} (a) Gate. (b) Progress $\bar t$. (c) Slope and signal ratio. Details: \textit{Supp.~E.1}.}
\label{fig:anneal}
\end{figure}

In Fig.~\ref{fig:anneal}, the gate decreases from about 0.45 to 0.12 as the mean progress $\bar t$ rises from about 0.40 to 0.82; Eq.~\eqref{eq:msa_closed_form} with $\varepsilon_d\approx\varepsilon_J=0.25$ gives $r(\bar t)-r^+\approx 0.48$ and $0.10$.
The guidance thus shrinks with the remaining error (Prop.~\ref{prop:residual}) and stays on while predictions lag; the mean push is 0.2 to 0.4 of the base field, below the clip $\chi=1$.
MSA has the highest slope among the plotted methods at every checkpoint, from 10k (about 0.25 vs.\ 0.08 to 0.16) to 100k (0.746 vs.\ at most 0.713).

\subsection{Comparison with UW-Flow}
\label{sec:exp:uwflow}

Table~\ref{tab:uwflow} adds UW-Flow~\cite{jeon2026uwflow}, which relaxes both marginals of every transport problem.
Relaxing the attraction to $y^+$ spreads the cloud: UW-Flow has over twice the noise of W-Flow (7.80 vs.\ 3.30; 5.15 vs.\ 2.33~px).
With accentuation, a contraction from the relaxed transport to $o_t$ opposes this spreading; at 15k the two nearly cancel (\textit{Supp.~D.1 and~D.2}), and UW-Flow + acc.\ has the best $\tau$ (.320 vs.\ .315 for MSA).
Its over-dispersion at 40k ($\tau$ .177, tail share .542) is consistent with the loss of this balance.
MSA keeps the base transport balanced and adds one shared translation in closed form, so it has no such term: it is stable (tail share .000), with the best slope and EPE at 15k and the best $\tau$ and slope at 40k; its noise is not the lowest.

\section{Conclusion}
\label{sec:conclusion}

We introduced Mass-Share Accentuation (MSA) to remove the target bias of action accentuation in drifting world models, whose push persists at the recorded future and, in kernel and balanced fields, settles beyond it.
By subtracting the correct future's share, MSA's guidance vanishes at the target, as a controlled toy confirms.
On Push-T, Robomimic, and BridgeV2, where models under-react, MSA improves action response and motion fidelity over latent-space baselines at no inference cost.

\clearpage 

\section*{LLM Usage Statement}
LLMs assisted with drafting, editing, LaTeX formatting, and consistency checks of the text. All ideas, derivations, experiments, and results are the authors'; all assisted text was verified by the authors.

{
    \small
    \bibliographystyle{ieeenat_fullname}
    \bibliography{main}
}


\end{document}